\documentclass[11pt,a4paper]{article}
\usepackage{times,latexsym}
\usepackage{url}
\usepackage[T1]{fontenc}
\usepackage{amssymb}
\usepackage{booktabs}

\usepackage[acceptedWithA]{tacl2021v1}
\usepackage{graphicx}
\usepackage{amsmath}
\usepackage{physics2}
\usephysicsmodule{ab}
\usepackage{enumerate}
\usepackage{xspace,mfirstuc,tabulary}

\newif\iftaclinstructions
\taclinstructionsfalse % AUTHORS: do NOT set this to true
\iftaclinstructions
\renewcommand{\confidential}{}
\renewcommand{\anonsubtext}{(No author info supplied here, for consistency with
TACL-submission anonymization requirements)}
\newcommand{\instr}
\fi

\iftaclpubformat % this "if" is set by the choice of options

\else

\fi

\title{Rethinking the Relationship between the Power Law and Hierarchical Structures}

\author{
Kai Nakaishi$^{1, 2, \ast}$
   \
   Ryo Yoshida$^3$
   \
   Kohei Kajikawa$^{4, \ast}$
   \
   Koji Hukushima$^3$
   \
   Yohei Oseki$^{3, 5}$
   \\
   $^1$RIKEN, Japan 
   \
   $^2$National Institute for Japanese Language and Linguistics, Japan
   \\
   $^3$The University of Tokyo, Japan
   \
   $^4$Georgetown University, USA
   \\
   $^5$National Institute of Informatics, Japan
   \\
   \texttt{kai.nakaishi@riken.jp}
   \\
   \texttt{\{yoshiryo0617, k-hukushima, oseki\}@g.ecc.u-tokyo.ac.jp}
   \\
   \texttt{kk1571@georgetown.edu}
 }

\date{}

\begin{document}
\maketitle

\renewcommand{\thefootnote}{\fnsymbol{footnote}}
\footnotetext[1]{This work was conducted while KN and KK were at the University of Tokyo.}

\begin{abstract}
Statistical analysis of corpora provides an approach to quantitatively investigate natural languages. This approach has revealed that several power laws consistently emerge across different corpora and languages, suggesting universal mechanisms underlying languages. In particular, the power-law decay of correlations has been interpreted as evidence of underlying hierarchical structures in syntax, semantics, and discourse. This perspective has also been extended beyond corpora produced by human adults, including child speech, birdsong, and chimpanzee action sequences. However, the argument supporting this interpretation has not been empirically tested in natural languages. To address this gap, the present study examines the validity of the argument for syntactic structures. Specifically, we test whether the statistical properties of parse trees align with the assumptions in the argument. Using English and Japanese corpora, we analyze the mutual information, deviations from probabilistic context-free grammars (PCFGs), and other properties in natural language parse trees, as well as in the PCFG that approximates these parse trees. Our results indicate that the assumptions do not hold for syntactic structures and that it is difficult to apply the proposed argument not only to sentences by human adults but also to other domains, highlighting the need to reconsider the relationship between the power law and hierarchical structures.
\end{abstract}

\section{Introduction}
\label{sec:intro}

Natural languages exhibit characteristic statistical properties. Quantifying these properties through corpus-based analyses is crucial for understanding natural languages. Previous research has extensively explored this approach, uncovering various power laws, such as Zipf’s law~\citep{zipf2016human} and Heaps’ law~\citep{heaps1978information}, that consistently appear across different corpora and languages. These findings suggest the existence of universal mechanisms underlying these power laws.

The present study focuses on the power-law decay of correlation. Previous studies have measured correlations between elements such as phones \citep{sainburg2019parallels}, characters \citep{li1989mutual, ebeling1994entropy, ebeling1995long-range, lin2017critical, shen2019mutual}, and words \citep{tanaka2016long-range, takahashi2017neural, takahashi2019evaluating, mikhaylovskiy2023autocorrelations, nakaishi2024critical}, using metrics such as correlation functions and mutual information (MI), across various corpora. These studies have consistently found that the correlation decays according to a power-law function of the distance between elements. 

The power-law decay is qualitatively slower than the exponential decay, which is typically observed when interactions between elements are local, as in Markov chains. Therefore, the power-law decay of correlation in natural languages indicates that distant elements remain strongly correlated. This observation is consistent with linguistic intuition. For example, syntactic dependencies can, in principle, span arbitrarily long distances through repeated center-embedding, and a word introduced early in a context may still influence interpretation much later.

Moreover, the power-law decay of correlation has been interpreted as evidence for underlying hierarchical structures. In natural languages, previous research has linked this power-law behavior to linguistic hierarchies in syntax, semantics, and discourse~\citep{alvarez2006hierarchical, altmann2012origin, lin2017critical}. This interpretation has also been extended beyond speech and written texts by human adults. Similar power-law behavior has been reported in child speech \citep{sainburg2022long}, birdsong \citep{sainburg2019parallels, sainburg2021toward, youngblood2024language}, and chimpanzee action sequences \citep{howard2024nonadjacent}, leading to the claim that these may also possess some form of underlying hierarchical structures, which need not correspond to linguistic hierarchies such as syntax and semantics.

This interpretation is justified by \citet{lin2017critical}. They showed that the decay of correlation follows a power law in sequences generated by probabilistic context-free grammars (PCFGs; \citealp{jelinek1992basic}), which are a simple model for generating hierarchical tree structures. To be more precise, they proved that in PCFGs, the correlation decays exponentially in sufficiently large structures. In addition, they assumed that the sequential distance grows exponentially with the structural distance. These two give rise to the power-law decay of correlation in sequences.

However, several assumptions underlying this argument seem implausible in light of well-established linguistic facts. Yet, previous studies have not tested the validity of these assumptions quantitatively using empirical data.

To address this gap, we test whether these assumptions hold for hierarchical structures in natural language syntax, i.e., parse trees. Syntactic structures have been well formalized, resulting in the availability of large-scale treebanks. This enables statistical analyses to test these assumptions.

Specifically, we formalize three key questions essential for examining the validity of the assumptions: (i) Does the correlation decay exponentially in syntactic structures?; (ii) Does the sequential distance grow exponentially with the structural distance?; (iii) Do the statistical properties of syntactic structures deviate significantly from those of PCFGs? To answer these questions, we analyze English and Japanese treebanks.

Our findings are negative for all three questions, indicating that the assumptions in the proposed argument do not hold for syntactic structures in natural languages. Therefore, it is necessary to reconsider the relationship between the two facts, the power-law decay of correlation and the hierarchical syntactic structures in natural languages. 

In addition, we perform the same analyses on trees generated by the PCFG that approximates parse trees, to highlight differences between natural language parse trees and trees generated by the PCFG. These analyses further show that trees generated by the PCFG are too small to exhibit the asymptotic behavior assumed in the argument of \citet{lin2017critical}. This observation raises doubts about the applicability of their argument not only to single sentences by adults but also to other sequences whose lengths are comparable to those of single sentences, such as child speech, birdsong, and chimpanzee action sequences.

While the present study focuses on the hierarchical syntactic structures of individual sentences, the structures of entire texts consisting of multiple sentences, such as discourse structures, also appear to be hierarchical. The argument by \citet{lin2017critical}, if applied to these structures, might account for the power-law decay of correlation in natural languages. 

It would therefore be fruitful to investigate the relationship between the power-law behavior and hierarchical structures at the discourse level. Although previous studies have proposed formal frameworks for discourse annotation \citep{mann-thompson-1988,prasad-etal-2014-reflections,zeldes-etal-2025-erst}, the size of existing corpora remain too small to examine the validity of the argument. Developing sufficiently large-scale discourse corpora is an important next step for future research.

Another direction is to explore alternative mechanisms that may give rise to power-law behavior in natural languages. One such mechanism is \textit{critical phenomena} in the statistical physics literature.

\section{Background}

\subsection{Power-law Decay of Correlation}

Previous research has shown that the correlation in natural languages decays according to a power-law function of distance. This phenomenon has been observed across various corpora and languages \citep{li1989mutual, ebeling1994entropy, ebeling1995long-range, tanaka2016long-range, lin2017critical, takahashi2017neural, shen2019mutual, takahashi2019evaluating, sainburg2019parallels, mikhaylovskiy2023autocorrelations, nakaishi2024critical}.

The significance of the power-law decay of correlation is clear in comparison with the exponential decay. For example, in signals generated sequentially by a Markov chain, the correlation $C$ is 
asymptotically exponential with the distance $r$, i.e., $C \sim \exp \ab ( - \lambda r )$~\citep{li1990mutual, lin2017critical}. In this case, if the distance is sufficiently larger than $\xi = 1 / \lambda$, the correlation is negligible. Thus, $\xi$, which is called the \textit{correlation length} in statistical physics~\citep{stanley1987introduction}, means the scale within which the correlation is significant.

In contrast, the power-law decay is expressed as $C \sim r^{- \alpha}$. Although the right-hand side resembles a polynomial form, the exponent $\alpha$ is a real number rather than a natural number. This decay is slower than the exponential one for any correlation length $\xi$, meaning that the correlation is significant for any long distances. In other words, the scale of correlation diverges.

Moreover, power-law functions are scale-invariant. For the scale transformation $r' = k r$, we can rescale the correlation by $C' = k^{- \alpha} C$ so that the relation remains the same, $C' \sim r'^{- \alpha}$. This property also implies the divergence of the scale. 

Therefore, the power-law decay of correlation observed in various corpora means that the scale diverges in natural languages. Intuitively, this implies that a local change can propagate and affect the global structure of a text.

\subsection{Power Law and Hierarchical Structures}
\label{subsec:power_and_hierarchy}

In studies of human language, this power-law behavior has been interpreted as evidence of underlying hierarchical structures. This interpretation is based on the work of \citet{lin2017critical}, which shows that the decay of correlation follows a power law in sequences generated by PCFGs.

PCFGs generate tree structures in a top-down manner, from the root to the leaves iteratively, where the category of a parent node determines the categories of its children. The leaves correspond to a surface sequence, while the whole tree represents the underlying hierarchical structure.

PCFGs are a probabilistic extension of context-free grammars \citep{chomsky1956three}, which were originally introduced to describe the nested structures, such as the \textit{if…then} construction. PCFGs have also been used to model syntactic structures \citep{charniak1997Statistical}, due to their simplicity and ability to capture fundamental aspects of hierarchical syntactic structures.

\begin{figure}[b]
    \centering
    \includegraphics[width=\linewidth]{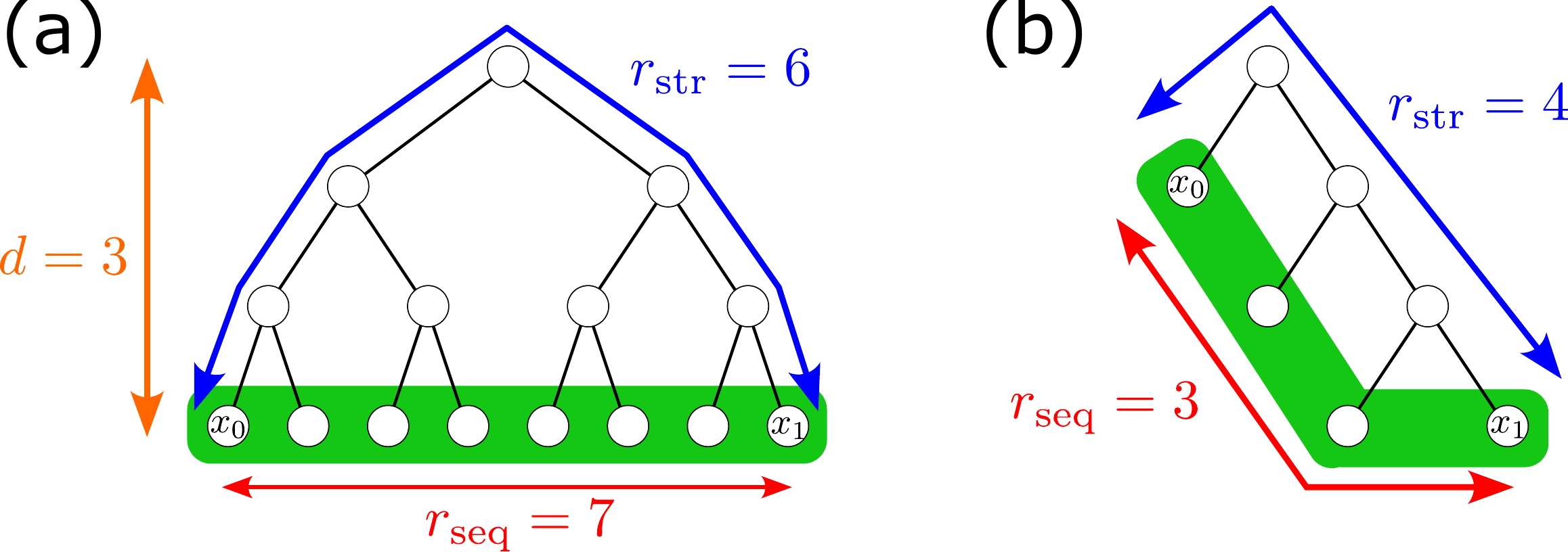}
    \caption{Sequential and structural distances. (a) If trees are balanced, the sequential distance grows exponentially with the structural distance, $r_{\text{seq}} \sim \exp ( \mu  r_{\text{str}})$. (b) If trees are strongly biased, the growth is slower. For example, the relation is linear, $r_{\text{seq}} \sim r_{\text{str}}$.}
    \label{fig:PCFG}
\end{figure}

We provide an overview of the discussion in \citet{lin2017critical}. To understand their discussion, it is useful to distinguish between \textit{sequential} and \textit{structural} distances, which we refer to as $r_{\text{seq}}$ and $r_{\text{str}}$, respectively. The former is defined along the sequence. The latter is the distance in the hierarchical structure behind the sequence, as shown in Figure~\ref{fig:PCFG}. 

\citet{lin2017critical} have proven that in PCFGs, the correlation decays exponentially with the structural distance, $C \sim \exp ( - \lambda r_{\text{str}} )$, at sufficiently large distances. Note that the decay may follow non-exponential forms at shorter distances. They also assumed that generated trees are balanced, i.e., the trees have neither left- nor right-branching bias, as in Figure~\ref{fig:PCFG}~(a). Letting the depth of the two nodes from the common ancestor be $d$, the structural distance is roughly $r_{\text{str}} \sim 2 d$, and the sequential distance is exponential with $d$, $r_{\text{seq}} \sim q^d$, where $q$ is the typical number of children of each node. Thus, the sequential distance grows exponentially with the structural distance, $r_{\text{seq}} \sim \exp (\mu r_{\text{str}})$, where $\mu = (\ln q) / 2$. Combining $C \sim \exp ( - \lambda r_{\text{str}} )$ and $r_{\text{seq}} \sim \exp (\mu r_{\text{str}})$ leads to the power-law decay of correlation in sequences, $C \sim r_{\text{seq}}^{- \alpha}$, with $\alpha = \lambda / \mu$. 

This discussion demonstrates that the correlation in sequences decays according to a power law if the sequences have hierarchical structures that satisfy the two properties: First, the correlation decays exponentially with the structural distance, $C \sim \exp \ab ( - \lambda r_{\text{str}} )$; second, the structures are balanced, so that the growth of sequential distance is exponential with the structural distance, $r_{\text{seq}} \sim \exp ( \mu r_{\text{str}})$.

The present study aims to examine the validity of this proposed argument. Throughout this study, we focus on syntactic structures. It is widely accepted in linguistics that these structures are hierarchical, with well-established formal descriptions. Using large-scale datasets based on this formalism, we can test if the proposed argument holds for syntactic structures.

It is worth noting that, in principle, PCFGs can be models not only for syntactic structures but also for more general hierarchical structures. Indeed, PCFGs have been applied to represent hierarchical structures in diverse domains \citep{ellis2015unsupervised, worth2005growing, gilbert2007probabilistic, knudsen1999rna,harlow2012tree, li1989spatial, li1991expansion, tano2020towards, lieck2021recursive}.

From this perspective, the argument by \citet{lin2017critical} may also apply to other types of linguistic structures, such as semantic and discourse ones. For example, entire texts composed of many sentences intuitively appear to have hierarchical structures defined by paragraphs, sections, chapters, and so on. This hierarchy may explain why the decay of correlation follows a power law beyond individual sentences, at the level of entire texts.

However, it is difficult to empirically test whether the argument holds for multi-sentence texts. Although previous studies have proposed formal frameworks for annotating discourse structures \citep{mann-thompson-1988, prasad-etal-2014-reflections, zeldes-etal-2025-erst}, existing discourse corpora are not yet large enough to reliably estimate the statistical quantities introduced in Section~\ref{sec:methods}, which we use to examine the validity of the assumptions in \citet{lin2017critical}.

\subsection{Rethinking Previous Research}
\label{subsec:rethinking}

To test whether the argument by \citet{lin2017critical} holds for syntactic structures in natural languages, several questions must be addressed.

First, do syntactic structures in natural languages actually exhibit the behaviors assumed in their argument? More specifically:
\begin{enumerate}[i)]
    \item Does the correlation in syntactic structures decay exponentially with the structural distance? 
    \item Are syntactic structures sufficiently balanced such that the sequential distance grows exponentially with the structural distance?
\end{enumerate}

\begin{figure}[t!]
    \centering
    \includegraphics[width=.9\linewidth]{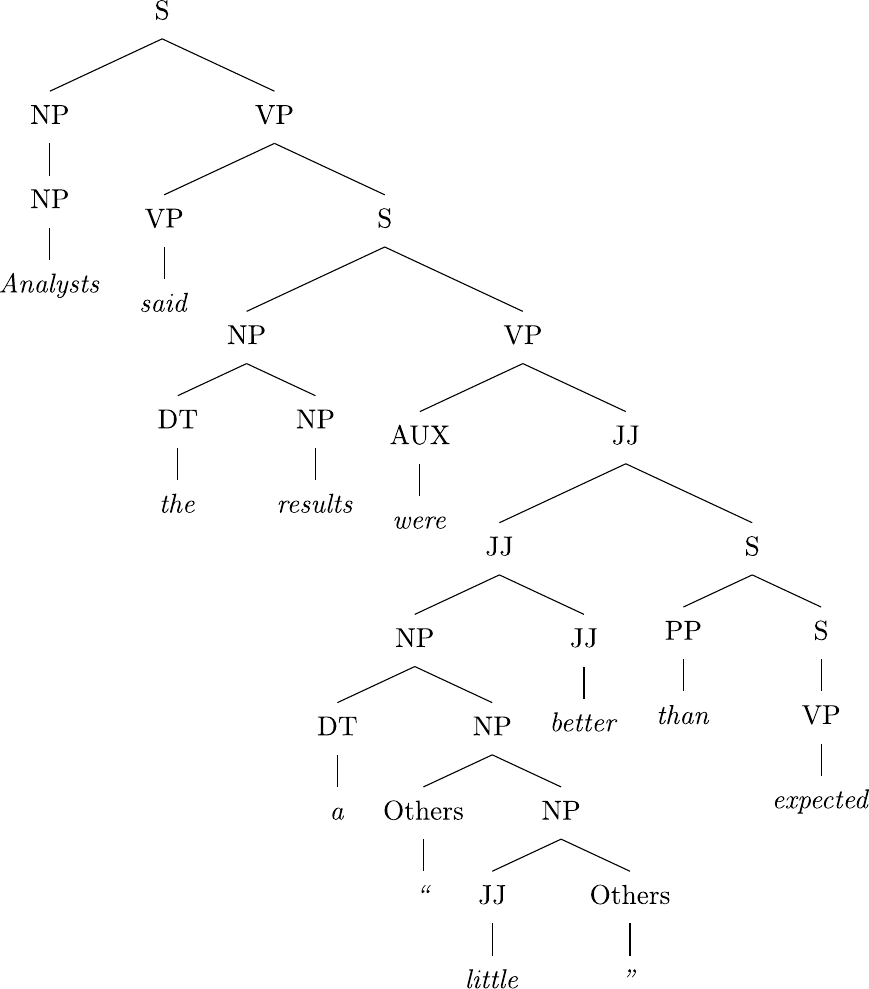}
    \caption{Example of a parse tree from BLLIP, preprocessed as described in Section~\ref{subsec_dataset}. This tree has a strong right-branching bias.}
    \label{fig:tree_example_BLLIP}
\end{figure}

In particular, the second one appears to contradict the well-known fact that syntactic structures typically exhibit a left- or right-branching bias~\citep{hawkins1990parsing, dryer1992greenbergian} (for instance, see Figure~\ref{fig:tree_example_BLLIP}). In such cases, the growth can be slower than exponential. For example, Figure~\ref{fig:PCFG}~(b) illustrates a case where the growth is linear, $r_{\text{seq}} \sim r_{\text{str}}$. 

In addition, \citet{lin2017critical} use PCFGs as models of hierarchical structures in natural languages. However, the expressive power of PCFGs is severely limited. In this model, the distribution of a subtree depends only on its root, a property we refer to as the \textit{context-free independence}. Previous studies have suggested that relaxing this property is necessary for modeling syntactic structures \citep{collins2003head, johnson2006adaptor, odonnell2009fragment}. Therefore, the last question is:
\begin{enumerate}[i)]
    \setcounter{enumi}{2}
    \item Do the statistical properties of syntactic structures deviate significantly from those of PCFGs?
\end{enumerate}

These three questions remain unresolved. First, although extensive research has measured correlations in texts and speech, the correlation in the underlying structures have not been quantified. The answer to question (i) remains elusive. Second, previous studies have not quantitatively examined how the branching bias affects the relationship between structural and sequential distances, which is crucial for addressing question (ii). Finally, it is well known that PCFGs cannot adequately model natural language syntax without being augmented to relax the context-free independence. However, to answer question (iii) clearly, we need to systematically quantify the differences in the statistical properties between natural language syntax and PCFGs.

In this study, we perform statistical analyses on parse trees to address these questions, one by one.

\section{Methods}
\label{sec:methods}

\subsection{Datasets and Preprocessing}
\label{subsec_dataset}

We primarily use the Brown Laboratory for Linguistic Information Processing 1987-89 Corpus Release 1 (BLLIP;~\citealp{charniak-etal-2000}). Its large scale enables precise statistical analyses.

To verify the robustness of our results, we also conduct the same analyses on two additional corpora. The first is WikiText \citep{merity-etal-2017}, an English corpus from a different domain than BLLIP. The second is the NINJAL Parsed Corpus of Modern Japanese (NPCMJ; \citealp{NPCMJ2016}), a treebank of Japanese, which is typologically distinct from English.

\paragraph{BLLIP}

BLLIP \citep{charniak-etal-2000} is a large-scale English treebank that provides Penn Treebank~\citep{marcus-etal-1993-building}-style syntactic annotations for three years' worth of Wall Street Journal articles. This treebank contains approximately 30 million words and one million sentences in total.
We remove the phonological null elements, such as traces of movement and ellipsis, to ensure alignment with WikiText-103~\citep{merity-etal-2017}, which does not include traces. 

The number of child nodes differs across nodes in the dataset, making it difficult to align nodes across different parse trees. To address this issue, we binarize the trees by recursively grouping child nodes from right to left, yielding right-branching structures.
As the binarization procedure introduces additional unary branches, we collapse all unary branches.
The binarization and unary-collapsing steps were implemented using the \texttt{Tree.chomsky\_normal\_form} and \texttt{Tree.collapse\_unary} methods in the NLTK library~\citep{bird-loper-2004-nltk}.
To provide intuition about how much these procedures affect tree shapes, we report depth, branching factor, and the proportion of unary nodes of the original trees in Appendix~\ref{app:treestats}.

In addition, we reduce the number of categories, since a larger category set requires more data for reliable statistical estimation. In particular, as categories become more fine-grained, correlations decrease and become difficult to distinguish from noise. Coarse-graining categories is crucial for capturing correlations.

Specifically, we replace part-of-speech (POS) and phrasal tags with a smaller set of categories. We refer to the Universal Dependencies~\citep{de-marneffe-etal-2021-universal}, which contains fewer types of tags than the Penn Treebank-style corpus, and group tags that tend to have similar types of dependents and heads. The eleven categories used in this study are listed in Appendix~\ref{app:tag}. We apply the same categories to both POS and phrasal tags, as \citet{lin2017critical} treat leaves and internal nodes in the same manner. An example of a preprocessed parse tree is shown in Figure~\ref{fig:tree_example_BLLIP}. 

The statistical quantities analyzed here are defined for the distribution of trees and estimated from the empirical dataset. Because parse trees depend on the chosen annotation scheme, these quantities also depend on the scheme. Nevertheless, we expect that the main results, such as whether the decay is exponential or power-law, will remain unchanged unless the annotation scheme is too fine-grained to capture correlations.

To verify this expectation, we also perform the same analyses under two alternative settings: the unbinarized setting and the phrasal-tag-only setting, where we distinguish phrasal tags from POS tags and focus only on the former. Appendix~\ref{app:scheme} shows that the results for both settings are qualitatively similar to those in the main text, that is, the answers to questions (i), (ii), and (iii) are negative.

Because the argument of \citet{lin2017critical} is based on simple PCFGs, the coarse-grained annotation scheme considered here is sufficient to test the validity of their argument. At the same time, statistical analyses under more refined annotation schemes would be valuable not only for assessing the robustness of our results but also from a linguistic perspective. However, since finer-grained annotation schemes require more data for reliable estimation of statistical quantities, analyses under such schemes would require the construction of annotated corpora much larger than BLLIP. We leave this direction for future work.

\paragraph{WikiText}

To examine whether our conclusions hold beyond BLLIP, we perform the same analyses on WikiText-103 (\texttt{wikitext-103-raw-v1} split) \citep{merity-etal-2017}, which consists of English Wikipedia articles that preserves the original Wikipedia markup and contains over 100 million tokens and approximately 3.9 million sentences. We implement a rule-based detokenizer and segment sentences with a \textsc{spaCy} \texttt{sentencizer}~\citep{honnibal2020spacy}. Then, we extract sentences up to 40 words (resulting in approximately 3.4 million sentences) and apply the Berkeley Neural Parser (\texttt{benepar\_en3} model)~\citep{kitaev-klein-2018,kitaev-etal-2019} to them to obtain Penn Treebank-style constituency trees. Finally, we binarize the trees and reduce the number of categories using the same method applied to BLLIP.

\paragraph{NPCMJ}

We also analyze the Japanese manually-annotated treebank, NPCMJ \citep{NPCMJ2016}.
It contains more than 1.4 million tokens and approximately 90,000 sentences. We first remove all of the phonologically null tokens (null pronouns, traces, etc.) and extract sentences of up to 40 words, resulting in approximately 82,000 sentences. We then binarize all parse trees and reduce the number of categories by systematically rewriting the POS and phrasal tags using a rule-based mapping scheme listed in Appendix~\ref{app:tag}.
Since Japanese is a head-final langauge, all trees are binarized in a left-branching direction, in contrast to the right-branching binarization applied to English.

As with BLLIP, Appendix~\ref{app:treestats} reports depth, branching factor, and the proportion of unary nodes for the original trees in NPCMJ, serving as a reference for assessing both the effect of preprocessing and the differences between English and Japanese. In addition, Appendix~\ref{app:scheme} reports results for unbinarized trees. The results are qualitatively similar to those for binarized trees, as in BLLIP, supporting the robustness of our conclusions.

\subsection{Mutual Information}
\label{sec:mi_def}

As a metric for correlation, we use the mutual information (MI) as in \citet{lin2017critical}. The MI between two tags is defined by
\begin{equation}
    I = \sum_{x_0, x_1} P(x_0, x_1) \ln \frac{P(x_0, x_1)}{P(x_0) P(x_1)},
    \label{eq:MI}
\end{equation}
where $P(x_0)$ and $P(x_1)$ are the marginal probabilities of the first and second tags, and $P(x_0, x_1)$ is their joint probability. The MI can be decomposed into
\begin{align}
    I 
    &= S_0 + S_1 - S_{01},
    \label{eq:MI_decompose}
\end{align}
where $S_0$, $S_1$, and $S_{01}$ are the Shannon entropy, defined by $S_0 = - \sum_{x_0} P(x_0) \ln P(x_0)$, $S_1 = - \sum_{x_1} P(x_1) \ln P(x_1)$, and $S_{01} = - \sum_{x_0, x_1} P(x_0, x_1) \ln P(x_0, x_1)$. By estimating $S_0$, $S_1$, and $S_{01}$, we can estimate the MI.

A naive estimate can be obtained by substituting the empirical distributions $\hat{P} (x_0)$, $\hat{P} (x_1)$, and $\hat{P} (x_0, x_1)$ into $P(x_0)$, $P (x_1)$, and $P(x_0, x_1)$ in the definitions of entropy. In this case, the estimates of entropy and MI are biased due to finite samples \citep{li1990mutual, arora2022estimating}.

To reduce this bias, we employ the more sophisticated estimator proposed by \citet{grassberger2003entropy}. In this method, the entropy $S_0$ is estimated by 
\begin{equation}
    \hat{S}_0
    = \psi (N) - \frac{1}{N} \sum_{x_0} n_{x_0} \psi (n_{x_0}),
    \label{eq:Grass}
\end{equation}
where $\psi$ is the digamma function, $n_{x_0}$ is the number of samples such that the former tag is $x_0$, and $N = \sum_{x_0} n_{x_0}$ is the total number of samples. $S_1$ and $S_{01}$ are estimated in a similar manner. \citet{grassberger2003entropy} has proven that the bias of this estimator rapidly converges to zero as the number of samples, $N$, increases. Therefore, if the estimated value saturates with respect to $N$, the finite-sample bias should be negligible.

To estimate the MI in sequences, we sample $N_{\text{data}}$ pairs of POS nodes separated by a sequential distance $r_{\text{seq}}$, from the dataset. For the MI in structures, we sample $N_{\text{data}}$ pairs of nodes at a fixed structural distance $r_{\text{str}}$, where the nodes are either POS or phrasal. We treat ordered pairs as distinct to avoid arbitrarily selecting one of the two. For example, we count pairs (AUX, PP) and (PP, AUX), corresponding to \textit{were} and \textit{than}, in Figure~\ref{fig:tree_example_BLLIP} as distinct to each other if both pairs are sampled.

In the following analyses, we compute the MI estimates while varying the number of samples, $N_{\text{data}}$. In our results, the estimates saturate as $N_{\text{data}}$ increases, indicating that the bias is sufficiently small.

To further support our claim, we also report the results obtained with six alternative entropy estimators \cite{madow1948limiting, miller1955note, zahl1977jackknifing, horvitz1952generalization, chao2003nonparametric, wolpert1995estimating, nemenman2001entropy} in Appendix~\ref{app:estimators}. The results are consistent with those reported in the main text, except for the estimators that were previously reported to exhibit larger errors \citep{arora2022estimating}.

\subsection{Growth of Sequential Distance}

To address question (ii), we investigate how the sequential distance depends on the structural distance. Specifically, for each $r_{\text{seq}}$ and $r_{\text{str}}$, we count the number of POS-tagged node pairs over the whole treebank whose sequential and structural distances are $r_{\text{seq}}$ and $r_{\text{str}}$, respectively. We also compute the average of sequential distance for each structural distance $r_{\text{str}}$.

\subsection{Context-free Independence Breaking}
\label{subsec:CFIB}

To address question (iii), it is necessary to quantify the extent to which syntactic structures deviate from PCFGs. We employ a metric for this quantification, called   \textit{context-free independence breaking} (CFIB; \citealp{nakaishi2024statistical}). 

The essential property of PCFGs, which we refer to as the \textit{context-free independence}, is that the distribution of categories of each node depends only on its parent. Consequently, given the categories of two nodes, their children are mutually independent. 

\begin{figure}[b]
    \centering
    \includegraphics[width=0.6\linewidth]{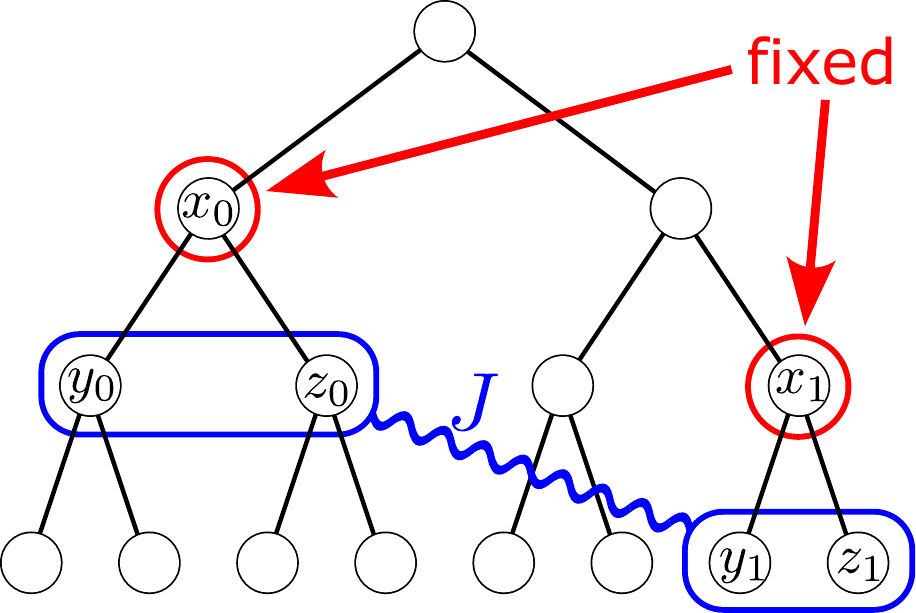}
    \caption{CFIB $J$ is the MI between the children of two nodes whose categories are fixed.}
    \label{fig:CFIB}
\end{figure}

Therefore, we can quantify the deviation of the distribution of trees from PCFGs by the MI between two pairs under the condition that the categories of their parents are fixed, as presented in Figure~\ref{fig:CFIB}. Specifically, provided that the categories of two nodes are $x_0$ and $x_1$, the CFIB is defined as 
\begin{equation}
    \begin{split}
        J_{x_0 x_1}
        = \sum_{y_0, z_0, z_0, z_1} 
        P( y_0, z_0, y_1, z_1 | x_0, x_1 ) \\
        \times \ln \frac{
            P( y_0, z_0, y_1, z_1 | x_0, x_1 )
        }{
            P( y_0, z_0 | x_0, x_1 )
            P( y_1, z_1 | x_0, x_1 )
        }.
    \end{split}
\end{equation}
Here, $P( y_0, z_0 | x_0, x_1 )$ denotes the conditional probability that the left and right children of the node with label $x_1$ are $y_0$ and $z_0$, respectively. Similarly, $P( y_1, z_1 | x_0, x_1 )$ refers to those for the node with label $x_1$. $P( y_0, z_0, y_1, z_1 | x_0, x_1 )$ is the conditional joint probability of both pairs of children. By definition, the CFIB is non-negative for any tree structures and zero for trees generated by PCFGs.

The CFIB is defined for the distribution of trees, as is the MI. We can estimate this metric from the dataset of trees, similarly to the estimation of the MI. We sample $N_{\text{data}}$ pairs of nodes that are tagged as $x_0$ and $x_1$, with the structural distance of $r_{\text{str}}$. Then, we compute the empirical distributions of tags on the children of the nodes. From these distributions, we estimate $J$ in the same procedure explained in Section~\ref{sec:mi_def}. We also distinguish between pairs and their permutations, as in the estimation of MI.

We verify the accuracy of the estimates for the CFIB by varying $N_{\text{data}}$, as we did for the MI. Appendix~\ref{app:estimators} also presents the results obtained with alternative entropy estimators, further supporting the validity of our estimates.

\subsection{PCFG Approximation}

We also construct a PCFG that approximates the distribution of natural language parse trees. This allows for a more detailed comparison between syntactic structures and PCFGs. The PCFG is constructed using the maximum likelihood estimation, which is equivalent to simply counting the frequency of each production rule in the dataset. Using the PCFG derived from the preprocessed dataset from BLLIP, we generate trees $1.6 \times 10^7$ times, with up to $100$ iterations per generation. This procedure yields approximately 13 million terminated trees. We then apply the same analyses to these generated trees as to parse trees. 

\subsection{Fitting}

We characterize the decay of correlation by fitting the data with exponential and power-law functions. Our goal is not to fit the data as accurately as possible or to estimate parameter values, but to test whether the observed behavior is closer to a power law or to an exponential form. For this purpose, using simpler models with fewer parameters is more appropriate, as it avoids issues such as overfitting. Accordingly, we use the two simple models, $ A\exp (- \lambda r )$ and $B r^{- \alpha}$.

To perform the fitting, we apply the least squares method after log-scaling the MI and CFIB values, because fitting with a linear scale puts too much weight on data points at shorter distances.

In model selection, model complexity should be taken into account, because goodness of fit, such as likelihood or squared error, trivially improves as the number of parameters increases. However, in our case, both models have two parameters. Therefore, it is sufficient to compare goodness of fit. Specifically, we use the reduced chi-square statistic, $\chi^2_{\nu}$. Under the commonly used assumption of Gaussian noise, comparing this statistic is equivalent to using standard model selection criteria such as the Akaike information criterion \cite{akaike1974aic} and the Bayesian information criterion \cite{schwarz1978estimating}.

We also fit the growth of sequential distance using an exponential function, $ C\exp ( \mu r )$, and a power-law function, $D r^{ \beta}$, without log-scaling the values.

For fitting, we use the \texttt{lmfit} Python package \cite{newville2014lmfit}. In the following results, we present the fitted parameters $A$, $\lambda$, and others along with the reduced chi-square statistic, $\chi^2_{\nu}$.

\subsection{Synthetic Data}

To demonstrate that our method can reliably estimate MI and distinguish between exponential and power-law decays, we apply it to mathematical models in which MI can be computed exactly. We generate random variables $X$ and $Y$ with joint probabilities defined as $P(X=0, Y=0) = P(X = 1, Y = 1) = (1 + \delta)/4$ and $P(X=0, Y=1) = P(X = 1, Y = 0) = (1 - \delta)/4$. The MI between $X$ and $Y$ is $I = \frac{1 + \delta}{2} \ln (1 + \delta) + \frac{1 - \delta}{2} \ln (1 - \delta)$, which is $\delta^2 / 2 + \mathcal{O}(\delta^3)$ as $\delta \to 0$. If $\delta = \exp (- \lambda r / 2)$, the MI decays asymptotically as $I \sim \exp(-\lambda r)$. If $\delta = r^{- \alpha / 2}$, the decay follows a power law, $I \sim r^{- \alpha}$. We refer to these as the exponential and power-law models, respectively.

\begin{figure*}[t!]
    \centering
    \includegraphics[width=.85\linewidth]{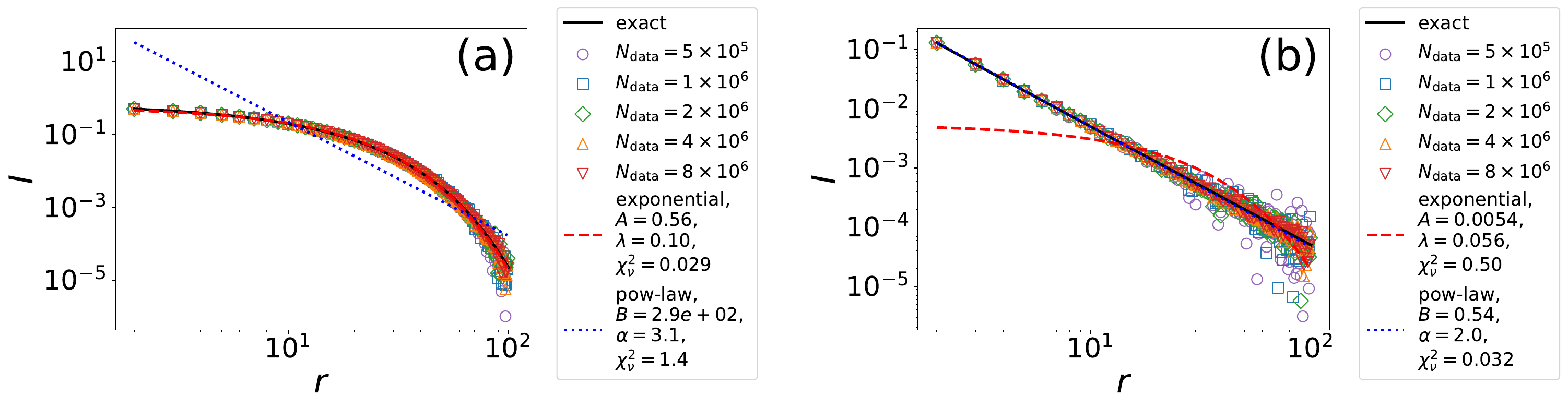}
    \caption{Estimation and fitting of the MI for (a) the exponential model with $\lambda = 0.1$ and (b) the power-law model with $\alpha = 2$. Fitting was performed for $N_{\text{data}} = 8 \times 10^6$.}
    \label{fig:result_toy}
\end{figure*}

We sample $X$ and $Y$ from the exponential model with $\lambda = 0.1$ and from the power-law model with $\alpha = 2$. We then estimate the MI and fit the resulting values. The results are shown in Figure~\ref{fig:result_toy}. In both cases, the estimates converge to the exact values as the sample size $N_{\text{data}}$ increases. In addition, $\chi^2_{\nu}$ is significantly smaller when the fitting function matches the model. Our method successfully discriminates between the exponential and power-law models.

\section{Results}

\subsection{Mutual Information in Sequences}
\label{subsec:MI_in_seq}

\begin{figure}[t!]
    \centering
    \includegraphics[width=.9\linewidth]{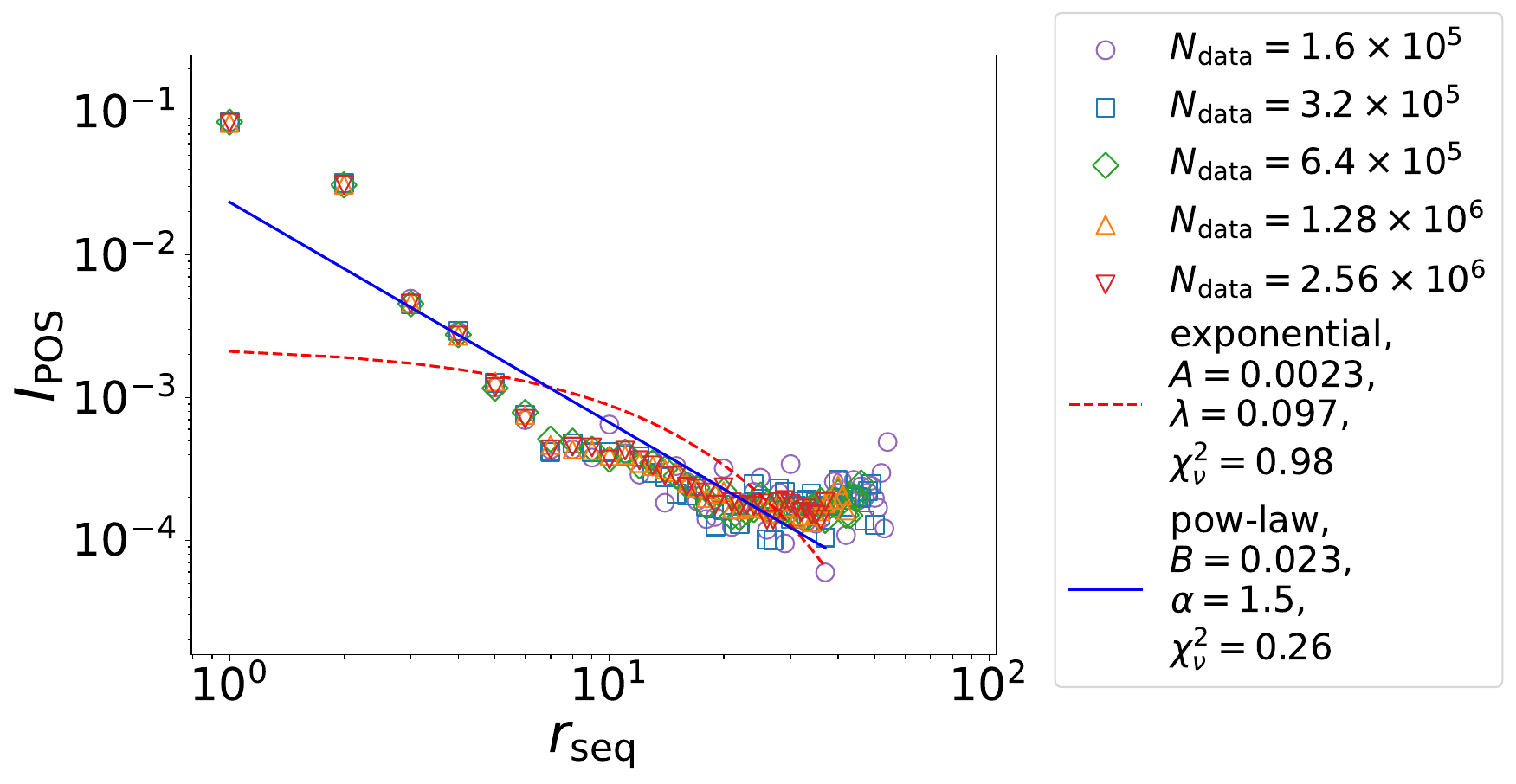}
    \caption{MI $I_{\text{POS}}$ between POS tags as a function of the sequential distance $r_{\text{seq}}$. The fitted exponential and power-law decays for $N_{\text{data}} = 2.56 \times 10^6$ are also presented. }
    \label{fig:result_IPOS}
\end{figure}

First, we confirm that the correlation in sequences decays according to a power law. Figure~\ref{fig:result_IPOS} shows the MI $I_{\text{POS}}$ between POS tags as a function of the sequential distance $r_{\text{seq}}$ for different data sizes $N_{\text{data}}$. The estimated values converge as $N_{\text{data}}$ increases, indicating that the estimation bias is sufficiently small. 

The decay of $I_{\text{POS}}$ is better fitted by a power-law function ($\chi^2_{\nu}=2.6 \times 10^{-1}$) than by an exponential one ($\chi^2_{\nu}=9.8 \times 10^{-1}$). This result indicates that the correlation between POS tags follows a power law, consistent with previous research.

\subsection{Mutual Information in Structures}
\label{subsec:MI_in_str}

\begin{figure}[t!]
    \centering
    \includegraphics[width=.9\linewidth]{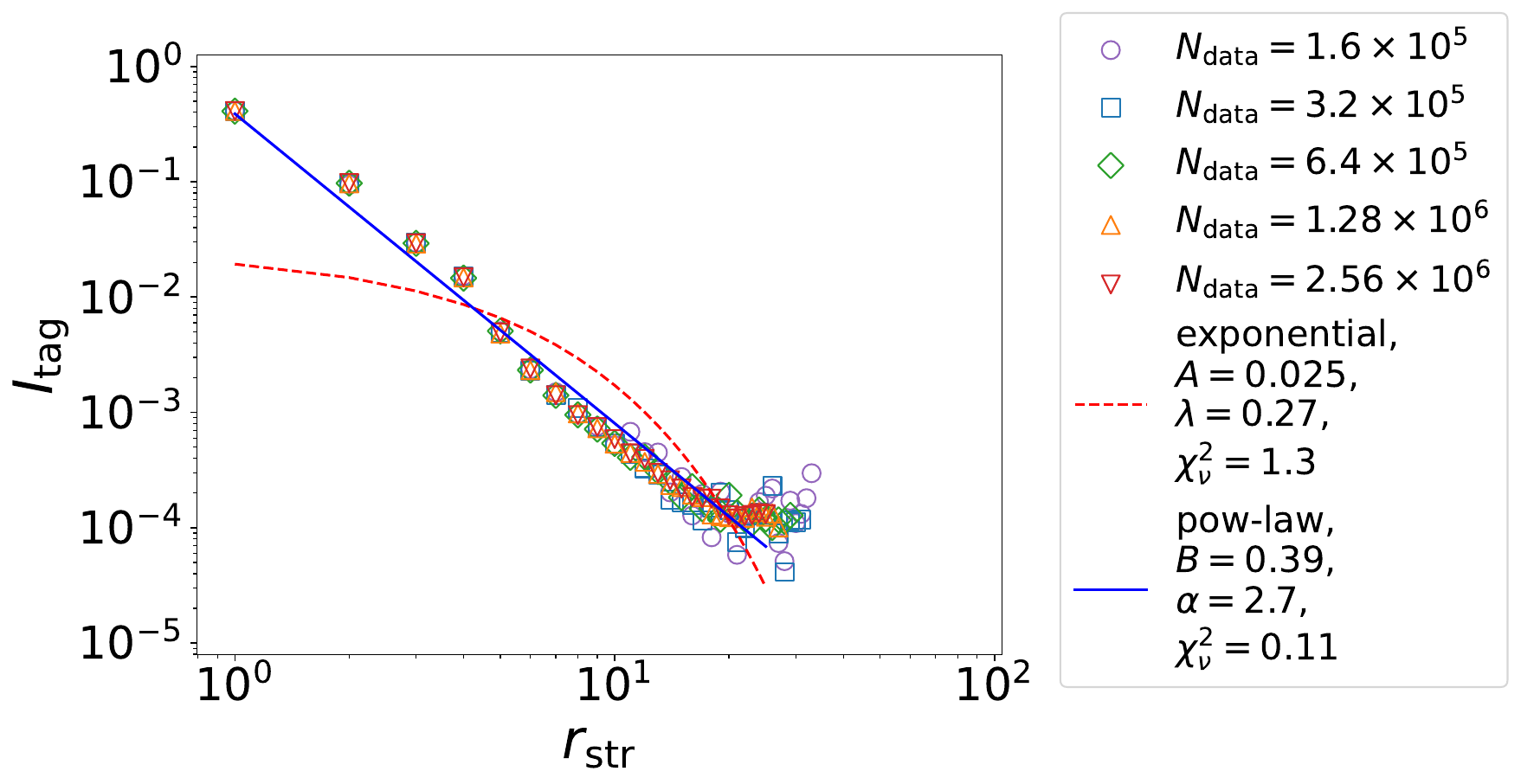}
    \caption{MI $I_{\text{tag}}$ between POS or phrasal tags as a function of the structural distance $r_{\text{str}}$. The fitted exponential and power-law decays for $N_{\text{data}} = 2.56 \times 10^6$ are also presented. }
    \label{fig:result_Itag}
\end{figure}

We now address question (i): whether the exponential decay of correlation with the structural distance holds in natural language syntactic structures. The estimated MI between POS or phrasal tags as a function of the structural distance is shown in Figure~\ref{fig:result_Itag}. As in the previous section, the convergence of the estimated values with increasing $N_{\text{data}}$ indicates that the bias is negligible.

Model fitting reveals that the correlation decays according to a power law ($\chi^2_{\nu}= 1.1 \times 10^{-1}$) rather than an exponential form ($\chi^2_{\nu}=1.3$) with respect to the structural distance. This result indicates that syntactic structures do not meet the condition necessary for the proposed argument to hold.

\subsection{Growth of Sequential Distance}
\label{subsec:seq_dist}

\begin{figure}[t!]
    \centering
    \includegraphics[width=.9\linewidth]{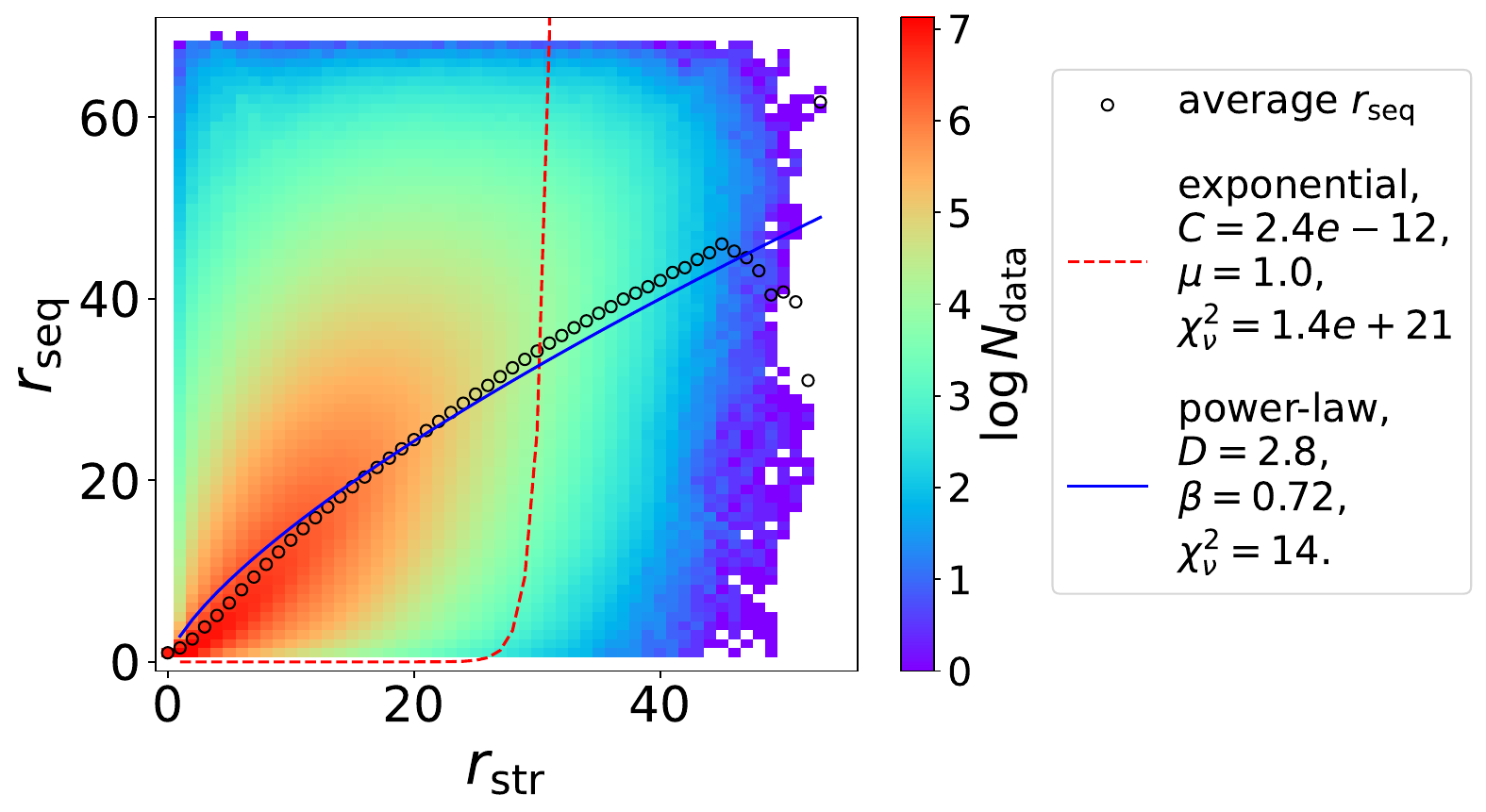}
    \caption{Number of POS tag pairs such that the structural and sequential distances are $r_{\text{str}}$ and $r_{\text{seq}}$, respectively. Black circles represent the average sequential distance for each structural distance.}
    \label{fig:result_r}
\end{figure}

Question (ii) asks whether the sequential distance grows exponentially with increasing structural distance. To examine this, we count the number of POS tag pairs such that the structural and sequential distances are $r_{\text{str}}$ and $r_{\text{seq}}$, respectively. The results are shown in Figure~\ref{fig:result_r}. We also compute the average sequential distance for each structural distance, represented by black circles, and fit the results using exponential and power-law functions.

The fitting reveals that the growth follows a power law ($\chi^2_{\nu}=1.4 \times 10$) rather than an exponential form ($\chi^2_{\nu}=1.4 \times 10^{21}$), although the trend becomes less clear for larger structural distances $r_{\text{str}} \gtrsim 50$ because of data sparsity. Notably, the exponent $\beta=0.72$ is smaller than $1$, indicating that the growth is slower than linear. As demonstrated in Figure~\ref{fig:PCFG}~(b), this slow growth quantitatively reflects a strong branching bias in syntactic structures. These findings clearly show that the second condition does not hold for syntactic structures.

\subsection{Context-free Independence Breaking}
\label{subsec:result_CFIB}

Question (iii) concerns whether the statistical properties of syntactic structures deviate significantly from those of PCFGs. We measure the CFIB $J$ as a metric of this deviation. Since reliable estimation of this metric requires a sufficiently large number of samples, we focus on the three most frequent cases, namely the fixed node pairs $(x_0, x_1) = (\text{NP}, \text{NP})$, $(\text{NP}, \text{VP})$, and $(\text{VP}, \text{VP})$.

\begin{figure}[t!]
    \centering
    \includegraphics[width=.9\linewidth]{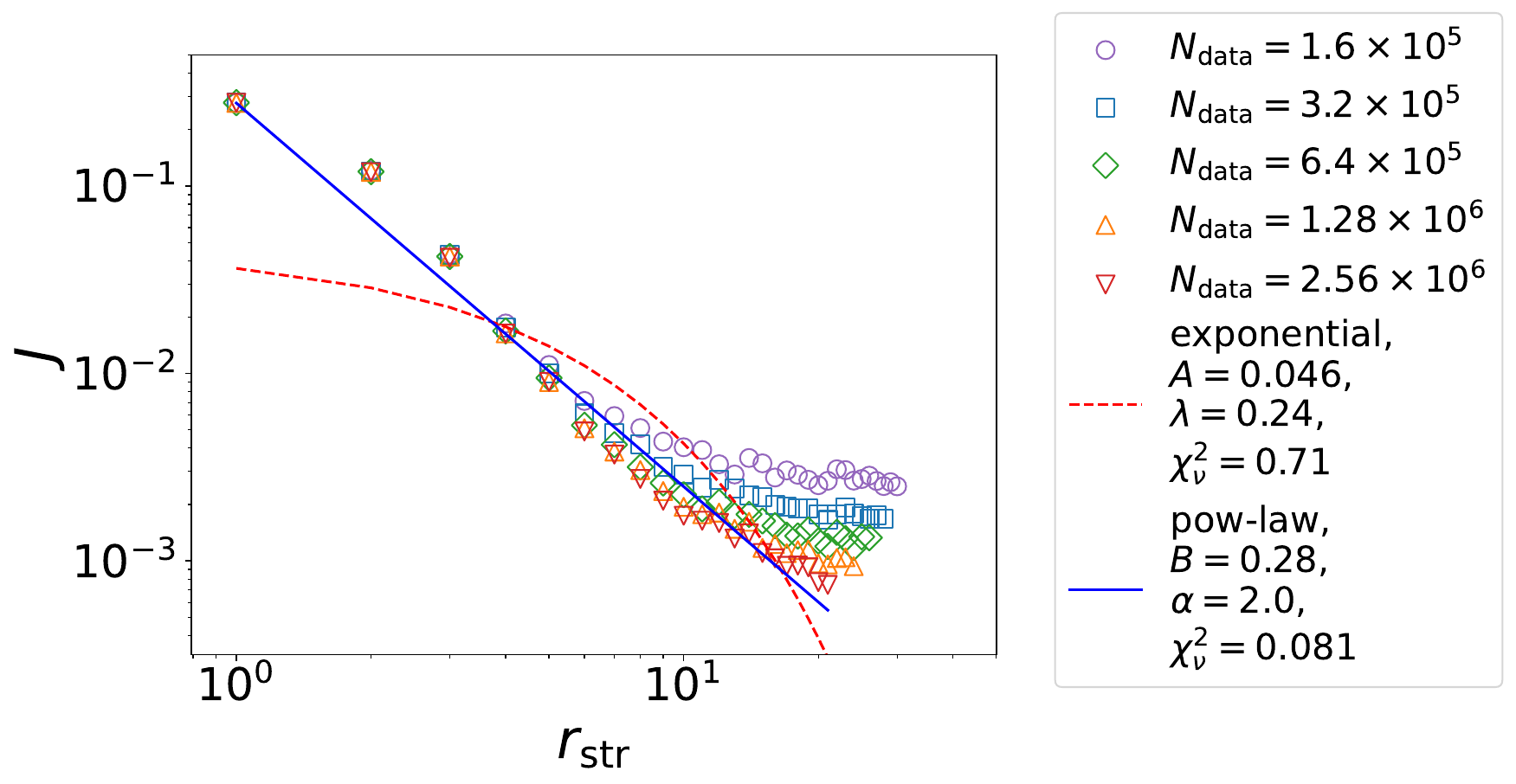}
    \caption{CFIB $J$ for $(x_0, x_1) = (\text{NP}, \text{NP})$ as a function of the structural distance $r_{\text{str}}$. The fitted exponential and power-law decays for $N_{\text{data}} = 2.56 \times 10^6$ are also presented.}
    \label{fig:result_J}
\end{figure}

The result for $(\text{NP}, \text{NP})$ is shown in Figure~\ref{fig:result_J}. The convergence of the estimates with increasing $N_{\text{data}}$ confirms that the estimation bias is negligible. The CFIB values are clearly positive, meaning that syntactic structures do not have the essential property of PCFGs. Furthermore, the metric decays according to a power law ($\chi^2_{\nu}=8.1 \times 10^{-2}$) rather than an exponential form ($\chi^2_{\nu}=7.1 \times 10^{-1}$) with respect to the structural distance, suggesting that the context-free independence is significantly broken even at large distances. These results indicate that the answer to question (iii) is negative.

This finding is also consistent with previous studies that developed PCFG-based parsers achieving high performance~\citep{collins2003head, johnson2006adaptor, odonnell2009fragment}. These studies augmented PCFGs to relax the context-free independence implicitly or explicitly.

As shown in Appendix~\ref{app:CFIB}, similar behaviors are observed for other frequent node pairs $(x_0, x_1) = (\text{NP}, \text{VP})$ and $(x_0, x_1) = (\text{VP}, \text{VP})$, suggesting that the observed deviation from PCFGs is not specific to a particular syntactic category pair.

\subsection{Other Corpora}
\label{subsec:other}

We examined whether the three assumptions in the argument of \citet{lin2017critical} hold for BLLIP. The results show that all of these assumptions are violated. The argument cannot account for the statistical properties of parse trees in this corpus. To test whether this conclusion is specific to BLLIP, we conduct the same analyses on WikiText and NPCMJ.

\begin{figure*}[t]
    \centering
    \includegraphics[width=\linewidth]{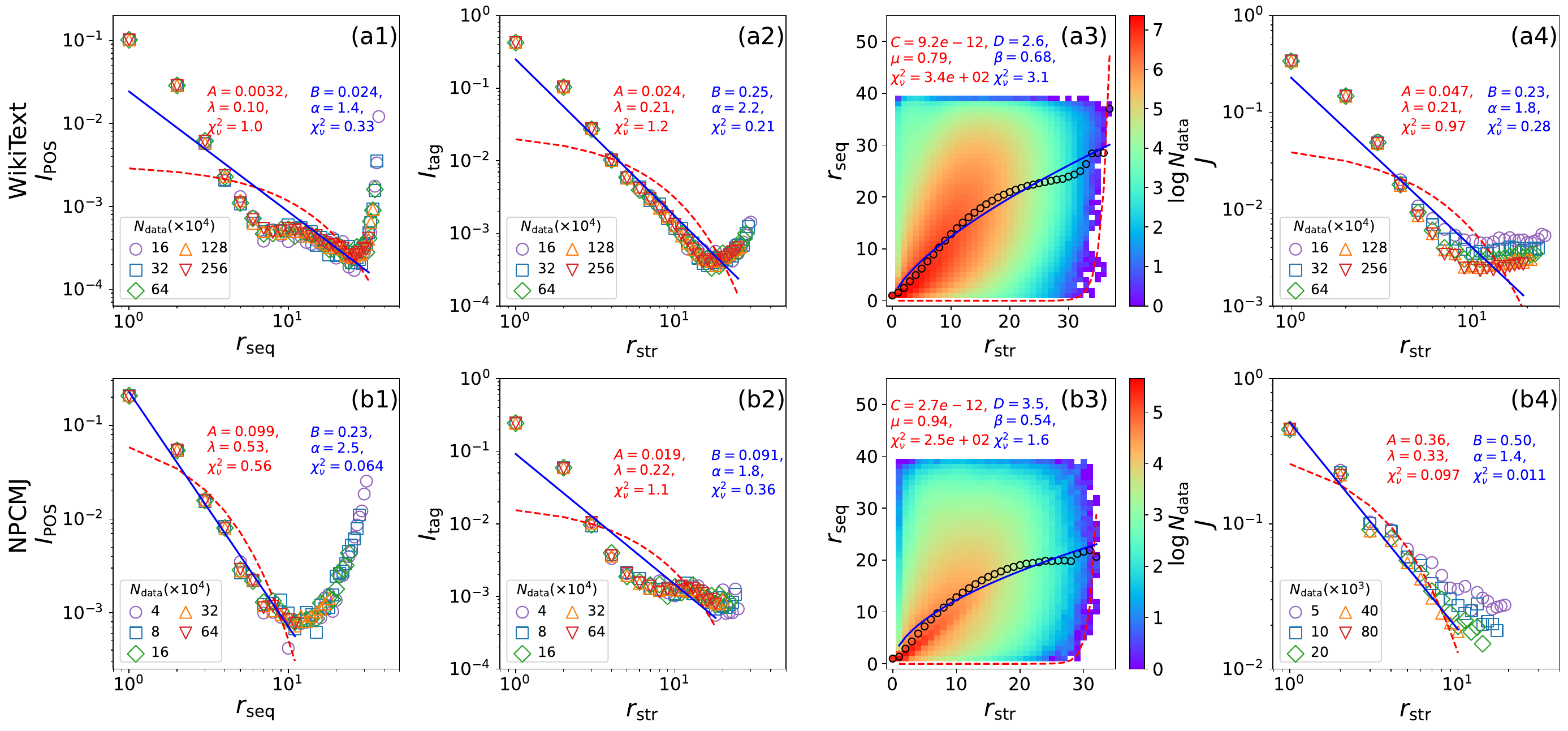}
    \caption{Statistical properties of syntactic structures in WikiText and NPCMJ. (a1--4) Results for WikiText: (a1) $I_{\text{POS}}$ as a function of $r_{\text{seq}}$; (a2) $I_{\text{tag}}$ as a function of $r_{\text{str}}$; (a3) relationship between $r_{\text{seq}}$ and $r_{\text{str}}$; and (a4) CFIB $J$ for $(x_0, x_1) = (\text{NP}, \text{NP})$ as a function of $r_{\text{str}}$. Fitting was performed with $N_{\text{data}} = 2.56 \times 10^6$. (b1--b4) Corresponding results for NPCMJ. Fitting was performed with $N_{\text{data}} = 6.4 \times 10^5$ for $I_{\text{POS}}$ and $I_{\text{tag}}$, and with $N_{\text{data}} = 4.0 \times 10^4$ for $J$.}
    \label{fig:result_other}
\end{figure*}

Figure~\ref{fig:result_other} summarizes the results. For both corpora, the statistical properties are similar to those observed for BLLIP: the MI decays according to a power law in both sequences and structures; the growth of sequential distance is slower than linear; and the CFIB also decays following a power law. These behaviors are supported by the reduced chi-square statistics, which are smaller for power-law fitting than for exponential fitting across all statistical quantities.

The results also exhibit slight differences from those for BLLIP. $I_{\text{POS}}$ and $I_{\text{tag}}$ for WikiText, as well as $I_{\text{POS}}$ for NPCMJ, increase at large distances. This effect may be attributable to structural differences in unusually long sentences or to potential parsing errors in such cases. However, we cannot determine the significance of this behavior because the datasets contain too few long sentences.

Nonetheless, the results clearly support that the assumptions proposed by \citet{lin2017critical} are violated not only for BLLIP, but also for WikiText and for NPCMJ. These results imply the robustness of our conclusion, given that Japanese is typologically distinct from English, exhibiting head-final, left-branching syntactic structures.

\subsection{PCFG Approximation}

\begin{figure*}[t]
    \centering
    \includegraphics[width=\linewidth]{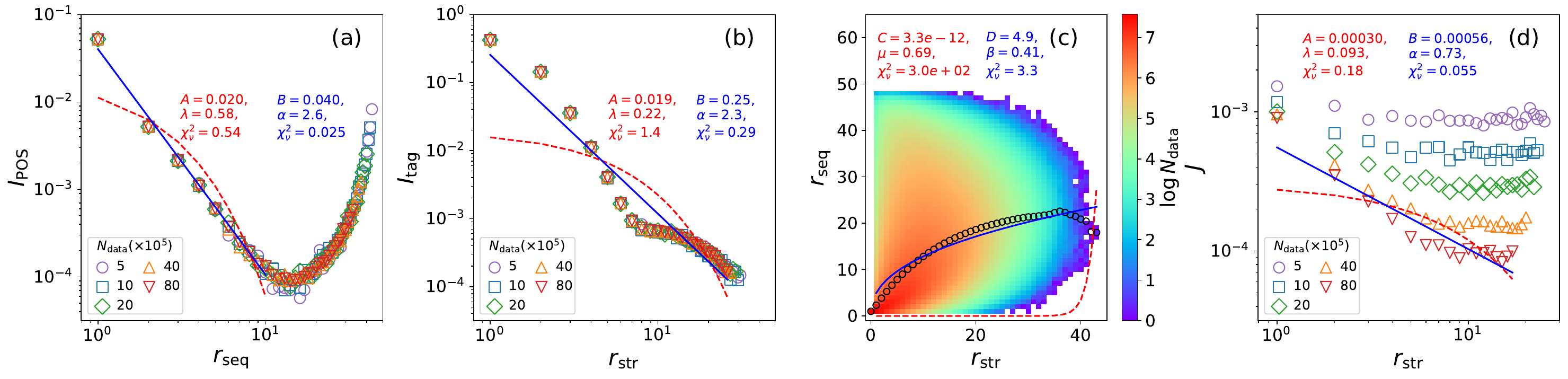}
    \caption{Statistical properties of trees generated by the PCFG obtained from the maximum likelihood estimation of syntactic structures. (a) MI $I_{\text{POS}}$ between POS tags. (b) MI $I_{\text{tag}}$ between POS or phrasal tags. (c) Distribution and average of sequential distances. (d) CFIB $J$ for $(x_0, x_1) = (\text{NP}, \text{NP})$. Fitting was performed for $N_{\text{data}} = 8 \times 10^6$ in (a) and (b), and for $r_{\text{str}} \leq 10$ in (a).}
    \label{fig:result_PCFG}
\end{figure*}

To further elucidate the difference between syntactic structures and PCFGs, we apply the same analyses to the PCFG constructed using maximum likelihood estimation on BLLIP. The results are summarized in Figure~\ref{fig:result_PCFG}.

\paragraph{Mutual Information in Sequences}

Figure~\ref{fig:result_PCFG}~(a) shows the MI in sequences. At short sequential distances, $r_{\text{seq}} \lesssim 10$, the MI decays according to a power law, although the decay rate does not clearly match that observed in parse trees. At larger distances, $r_{\text{seq}} \gtrsim 10$, the MI increases substantially, in contrast to the trend in parse trees.

\paragraph{Mutual Information in Structures}

The MI in structures, shown in Figure~\ref{fig:result_PCFG}~(b), exhibits a power-law decay similar to that in syntactic structures, although the decay rate differs. This result does not contradict the properties of PCFGs proven by \citet{lin2017critical}. While the MI in structures generated by PCFGs asymptotically follows an exponential decay at sufficiently large distances, the MI can exhibit different behavior at shorter distances. The non-exponential behavior of $I_{\text{tag}}$ suggests that for individual trees, the structural distances are too short for the asymptotic behavior to emerge.

This pre-asymptotic behavior in the PCFG constructed from parse trees raises the possibility that the statistical behavior of parse trees themselves is also pre-asymptotic. If so, it also calls into question the applicability of the argument of \citet{lin2017critical} to natural language parse trees, since the argument relies on asymptotic behavior.

This observation also has an important implication about the applicability of their argument to domains beyond texts and speech by human adults. As discussed in Section~\ref{sec:intro}, several studies have applied their argument to child speech and animal behavior, interpreting the power-law decay of correlations in these domains as evidence for the existence of some form of underlying hierarchical structures~\citep{sainburg2019parallels, sainburg2021toward, sainburg2022long, howard2024nonadjacent, youngblood2024language}.

However, in these domains, the sequence lengths are typically comparable to those in the PCFG studied here. In such cases, it is difficult to interpret the power-law decay of correlation as evidence for hierarchical structures, since the asymptotic behavior is unlikely to emerge.

\paragraph{Growth of Sequential Distance}

In Figure~\ref{fig:result_PCFG}~(c), the average sequential distance grows more slowly than exponentially and even than linearly, indicating that the generated trees are strongly biased. While this behavior is qualitatively similar to that in parse trees, the estimated exponent $\beta = 0.41$ significantly differs from that for the parse trees, where $\beta = 0.72$.

\paragraph{Context-free Independence Breaking}

The most striking difference between parse trees and trees generated by the PCFG appears in the CFIB metric. The true value of this metric is zero for PCFGs by definition. If the estimates converge to this true value with increasing data size, the corresponding values in the log-scale plot continue to decrease, approaching $-\infty$, since the logarithm of zero is $-\infty$.

In Figure~\ref{fig:result_PCFG}~(d), the estimates of CFIB is much smaller than those for syntactic structures in Figure~\ref{fig:result_J}. Furthermore, the estimates continue to decrease in the log-scale plot as $N_{\text{data}}$ increases, which is the expected behavior. These results are in contrast to those for parse trees.

\section{Discussion and Conclusion}

In natural languages, the correlation decays as a power law with respect to the sequential distance, as reported in numerous previous studies. At the same time, linguists generally agree that natural languages exhibit hierarchical syntactic structures. Although both phenomena are well established, their relationship remains elusive.

The present study examines this relationship, specifically the argument proposed by \citet{lin2017critical}. They attributed the power-law decay of correlation in natural languages to underlying hierarchical structures, which has since inspired numerous studies on both human languages and animal behavior. However, their argument relies on several assumptions that are questionable from a linguistic perspective. To examine the validity of the assumptions for syntactic structures, we test the three key questions through statistical analyses of parse trees in natural languages.

The analyses yield negative answers to all three questions: (i) The correlation in syntactic structures decays according to a power law with respect to the structural distance; (ii) The growth of the sequential distance is slower than linear with the structural distance, due to the strong branching bias in syntactic structures; (iii) The statistical properties of syntactic structures significantly differ from those of PCFGs. These results reveal that the relationship between the power-law decay of correlation and syntactic structures differs from that proposed by \citet{lin2017critical}.

We obtained similar results under various settings: for the English treebank BLLIP with several different schemes, for another English corpus, WikiText, and for the Japanese treebank NPCMJ. To further assess the robustness of our conclusions, it is important to examine statistical behavior under additional annotation schemes, domains, and languages. In particular, revealing the extent to which these behaviors are shared across languages is important from a linguistic perspective.

We also analyzed the PCFG obtained from the maximum likelihood estimation of BLLIP. The statistical properties of the PCFG clearly differ from those in parse trees. Furthermore, this analysis demonstrates that the asymptotic behavior of the PCFG does not emerge within individual sequences. This implies the difficulty of applying the proposed argument not only to sentences produced by human adults but also to child speech and animal behavior, where sequence lengths are comparable to those for the PCFG.

In addition, the statistical differences between natural language parse trees and trees generated by PCFGs may have implications for theoretical research on large language models. Some studies in this field analyze models trained on synthetic data, in which PCFGs are sometimes used to generate the data \citep{allen-zhu2023physics1, cagnetta2024towards}. However, the model’s ability to learn PCFGs will not account for its ability to learn syntactic structures of natural languages, although some studies employ PCFGs as models of more abstract structures than syntax.

It is noteworthy that the pre-asymptotic behavior in the PCFG constructed from parse trees does not serve for a definitive evidence against the validity of the argument by \citet{lin2017critical}, because this behavior does not necessarily mean that parse trees themselves also exhibit pre-asymptotic behavior. Our main evidence instead come from the direct analyses of parse trees in Sections~\ref{subsec:MI_in_seq}--\ref{subsec:other}, which clearly show that the statistical properties of natural language parse trees for their typical sizes differ from those assumed in the proposed argument.

Currently, we cannot rule out the possibility that the MI in natural language syntactic structures follows an exponential decay at structural distances much larger than those observed in this study. However, even if this were the case, it would not explain the power-law decay of correlation in natural language sequences, because such large parse trees would be too rare to have a significant influence on the statistical properties of the corpus. Moreover, as long as syntactic structures are biased, the sequential distance grows more slowly than exponentially, resulting in a decay of MI in sequences that is faster than a power law. Therefore, statistical analyses of larger parse trees are unlikely to account for the power-law decay of correlation observed in natural languages.

A more promising direction is to consider structures larger than individual sentences, namely, the hierarchical organization of entire texts composed of multiple sentences. The argument by \citet{lin2017critical} may account for the power-law decay of correlation observed across entire texts rather than within single sentences. As a first step in this direction, it will be necessary to develop sufficiently large-scale corpora annotated with discourse information to estimate statistical quantities such as $I_{\text{POS}}$, $I_{\text{tag}}$, and $J$.

Another direction is to explore approaches different from that of \citet{lin2017critical}. One alternative is \textit{critical phenomena} in statistical physics. These refer to the phenomena that statistical quantities exhibit power-law behaviors near \textit{phase transition} points, at which statistical properties of a system change qualitatively. Indeed, as the temperature parameter of large language models varies, generated texts exhibit critical phenomena, sharing statistical properties with natural languages, including the power-law decay of correlation~\citep{nakaishi2024critical}.

\section*{Acknowledgments}

We thank the action editor, Carlos G\'{o}mez-Rodr\'{i}guez, and the anonymous reviewers for their constructive and helpful feedback. We also thank Taiga Ishii, Shinnosuke Isono, Sho Yokoi, and Keisuke Sakaguchi for valuable discussions and suggestions. This work was supported by JSPS KAKENHI Grant Nos.~23H01095, 23KJ0622, 24H00087, and 25K24434; JST, PRESTO Grant No.~JPMJPR21C2; and the World-Leading Innovative Graduate Study Program for Advanced Basic Science Course at the University of Tokyo.

\bibliography{tacl2021}

@ARTICLE{akaike1974aic,
  author={Akaike, Hirotugu},
  journal={IEEE Transactions on Automatic Control}, 
  title={A new look at the statistical model identification}, 
  year={1974},
  volume={19},
  number={6},
  pages={716-723},
  doi={10.1109/TAC.1974.1100705}}

@article{allen-zhu2023physics1,
  title={Physics of Language Models: Part 1, Learning Hierarchical Language Structures},
  author={Allen-Zhu, Zeyuan and Li, Yuanzhi},
  journal={Available at SSRN},
  year={2023},
  doi={http://dx.doi.org/10.2139/ssrn.5250639},
}

@article{altmann2012origin,
  title={On the origin of long-range correlations in texts},
  author={Altmann, Eduardo G. and Cristadoro, Giampaolo and Esposti, Mirko Degli},
  journal={Proceedings of the National Academy of Sciences},
  volume={109},
  number={29},
  pages={11582--11587},
  year={2012},
  doi={10.1073/pnas.1117723109},
issn={1091-6490},
}

@article{alvarez2006hierarchical,
  title={Hierarchical structures induce long-range dynamical correlations in written texts},
  author={Alvarez-Lacalle, Enrique and Dorow, Beate and Eckmann, Jean-Pierre and Moses, Elisha},
  journal={Proceedings of the National Academy of Sciences},
  volume={103},
  number={21},
  pages={7956--7961},
  year={2006},
  doi={10.1073/pnas.0510673103},
issn={1091-6490},
}

@inproceedings{arora2022estimating,
    title = "Estimating the Entropy of Linguistic Distributions",
    author = "Arora, Aryaman  and
      Meister, Clara  and
      Cotterell, Ryan",
    editor = "Muresan, Smaranda  and
      Nakov, Preslav  and
      Villavicencio, Aline",
    booktitle = "Proceedings of the 60th Annual Meeting of the Association for Computational Linguistics (Volume 2: Short Papers)",
    month = may,
    year = "2022",
    address = "Dublin, Ireland",
    publisher = "Association for Computational Linguistics",
    doi = "10.18653/v1/2022.acl-short.20",
    pages = "175--195",
}

@inproceedings{
cagnetta2024towards,
title={Towards a theory of how the structure of language is acquired by deep neural networks},
author={Francesco Cagnetta and Matthieu Wyart},
booktitle={Advances in Neural Information Processing Systems 37},
year={2024},
doi={10.52202/079017-2645},
}

@article{chao2003nonparametric,
  title={Nonparametric estimation of Shannon’s index of diversity when there are unseen species in sample},
  author={Chao, Anne and Shen, Tsung-Jen},
  journal={Environmental and ecological statistics},
  volume={10},
  number={4},
  pages={429--443},
  year={2003},
  publisher={Springer},
  doi={10.1023/A:1026096204727},
}

@article{charniak1997Statistical,
  title={Statistical Techniques for Natural Language Parsing},
  author={Charniak, Eugene},
  journal={The AI Magazine},
  year={1997},
  volume={18},
  pages={33-44},
  doi = "10.1609/aimag.v18i4.1320",
issn = "2371-9621"
}

@article{chomsky1956three,
  title={Three models for the description of language},
  author={Chomsky, Noam},
  journal={IRE Transactions on information theory},
  volume={2},
  number={3},
  pages={113--124},
  year={1956},
  doi = {10.1109/TIT.1956.1056813},
issn = {0018-9448},
}

@article{collins2003head,
    title = "Head-Driven Statistical Models for Natural Language Parsing",
    author = "Collins, Michael",
    journal = "Computational Linguistics",
    volume = "29",
    number = "4",
    year = "2003",
    doi = "10.1162/089120103322753356",
    pages = "589--637"
}

@article{dryer1992greenbergian,
  title={The Greenbergian word order correlations},
  author={Dryer, Matthew S.},
  journal={Language},
  volume={68},
  number={1},
  pages={81-138},
  year={1992},
  doi = "10.2307/416370",
issn = "00978507"
}

@ARTICLE{ebeling1994entropy,
  title     = "Entropy and {Long-Range} Correlations in Literary English",
  author    = "Ebeling, Werner and P{\"o}schel, Thorsten",
  journal   = "Europhysics Letters",
  publisher = "IOP Publishing",
  volume    =  26,
  number    =  4,
  pages     = "241-246",
  month     =  may,
  year      =  1994,
issn = "1286-4854",
doi = "10.1209/0295-5075/26/4/001"
}

@ARTICLE{ebeling1995long-range,
  title     = "Long-range correlations between letters and sentences in texts",
  author    = "Ebeling, Werner and Neiman, Alexander",
  journal   = "Physica A",
  volume    =  215,
  number    =  3,
  pages     = "233-241",
  year      =  1995,
issn = "1873-2119",
doi = "10.1016/0378-4371(95)00025-3"
}

@inproceedings{ellis2015unsupervised,
 author = {Ellis, Kevin and Solar-Lezama, Armando and Tenenbaum, Josh},
 booktitle = {Advances in Neural Information Processing Systems 28},
 title = {Unsupervised Learning by Program Synthesis},
year = "2015"
}

@inproceedings{gilbert2007probabilistic,
  title={A Probabilistic Context-Free Grammar for Melodic Reduction?},
  author={{\'E}douard Gilbert and Darrell Conklin},
  year={2007},
booktitle = "Proceedings of the International Workshop on Artificial Intelligence and Music, 20th International Joint Conference on Artificial Intelligence",
}

@misc{grassberger2003entropy,
  author = {Grassberger, Peter},
  title  = {Entropy estimates from insufficient samplings},
  year   = {2003},
  note   = {{arXiv}:physics/0307138v2},
  doi = {10.48550/arXiv.physics/0307138},
}

@article{harlow2012tree,
  title={Tree-like structure of eternal inflation: A solvable model},
  author={Harlow, Daniel and Shenker, Stephen H. and Stanford, Douglas and Susskind, Leonard},
  journal={Physical Review D},
  volume={85},
  number={6},
  pages={063516},
  year={2012},
doi = "10.1103/PhysRevD.85.063516",
issn = "2470-0029"
}

@article{hawkins1990parsing,
  title={A parsing theory of word order universals},
  author={Hawkins, John A.},
  journal={Linguistic inquiry},
  volume={21},
  number={2},
  pages={223-261},
  year={1990},
  issn = "1530-9150"
}

@book{heaps1978information,
  title={Information Retrieval: Computational and Theoretical Aspects},
  author={Heaps, Harold Stanley},
  year={1978},
  publisher={Academic Press, Inc.}
}

@article{honnibal2020spacy,
author = {Honnibal, Matthew and Montani, Ines and Van Landeghem, Sofie and Boyd, Adriane},
doi = {10.5281/zenodo.1212303},
title = {{spaCy: Industrial-strength Natural Language Processing in Python}},
year = {2020}
}

@article{horvitz1952generalization,
  title={A generalization of sampling without replacement from a finite universe},
  author={Horvitz, Daniel G. and Thompson, Donovan J.},
  journal={Journal of the American statistical Association},
  volume={47},
  number={260},
  pages={663--685},
  year={1952},
  publisher={Taylor \& Francis},
  doi={10.1080/01621459.1952.10483446},
}

@article{howard2024nonadjacent,
  title={Nonadjacent dependencies and sequential structure of chimpanzee action during a natural tool-use task},
  author={Howard-Spink, Elliot and Hayashi, Misato and Matsuzawa, Tetsuro and Schofield, Daniel and Gruber, Thibaud and Biro, Dora},
  journal={PeerJ},
  volume={12},
  pages={e18484},
  year={2024},
doi = "10.7717/peerj.18484",
issn = "2167-8359",
}

@InProceedings{jelinek1992basic,
  author =  "Jelinek, Filip
and Lafferty, John D.
and Mercer, Robert L.",
  title =   "Basic Methods of Probabilistic Context Free Grammars",
  booktitle =   "Speech Recognition and Understanding",
  year =    "1992",
}

@inproceedings{johnson2006adaptor,
  author  = "Johnson, Mark and Griffiths, Thomas L. and Goldwater, Sharon",
  booktitle = {Advances in Neural Information Processing Systems 19},
  title   = "Adaptor grammars: A framework for specifying compositional nonparametric Bayesian models",
  year    =  2006,
}

@article{knudsen1999rna,
  title={{RNA} secondary structure prediction using stochastic context-free grammars and evolutionary history.},
  author={Knudsen, Bjarne and Hein, Jotun},
  journal={Bioinformatics},
  volume={15},
  number={6},
  pages={446-454},
  year={1999},
doi = "10.1093/bioinformatics/15.6.446",
issn = "1367-4811"
}

@TECHREPORT{li1989mutual,
  title       = "Mutual Information Functions of Natural Language Texts",
  author      = "Li, Wentian",
  institution = "Santa Fe Institute",
  year        =  "1989",
  number      = "89-10-008",
url = "https://www.santafe.edu/research/results/working-papers/mutual-information-functions-of-natural-language-t",
note        = {Accessed 2026-03-02},
}

@article{li1989spatial,
  title={Spatial 1/f spectra in open dynamical systems},
  author={Li, Wentian},
  journal={Europhysics Letters},
  volume={10},
  number={5},
  pages={395},
  year={1989},
doi = "10.1209/0295-5075/10/5/001",
issn = "1286-4854"
}

@article{li1990mutual,
  title={Mutual information functions versus correlation functions},
  author={Li, Wentian},
  journal={Journal of statistical physics},
  volume={60},
  pages={823-837},
  year={1990},
doi = "10.1007/BF01025996",
issn = "1572-9613"
}

@article{li1991expansion,
  title={Expansion-modification systems: a model for spatial 1/f spectra},
  author={Li, Wentian},
  journal={Physical Review A},
  volume={43},
  number={10},
  pages={5240},
  year={1991},
doi = "10.1103/PhysRevA.43.5240",
issn = "2469-9934"
}

@InProceedings{lieck2021recursive,
author="Worth, Peter and Stepney, Susan",
title="Recursive bayesian networks: Generalising and unifying probabilistic context-free grammars and dynamic bayesian networks",
booktitle="Advances in Neural Information Processing Systems 34",
year="2021",
pages="4370-4383",
}

@article{lin2017critical,
title = "Critical behavior in physics and probabilistic formal languages",
author = "Lin, Henry W. and Tegmark, Max",
journal = "Entropy",
year = "2017",
volume = "19",
number = "7",
pages = "299",
issn = "0885-2308",
doi = "10.3390/e19070299",
}

@article{madow1948limiting,
  title={On the limiting distributions of estimates based on samples from finite universes},
  author={Madow, William G.},
  journal={The Annals of Mathematical Statistics},
  pages={535--545},
  year={1948},
  publisher={JSTOR},
  doi={10.1214/aoms/1177730149},
}

@misc{mikhaylovskiy2023autocorrelations,
  author = {Mikhaylovskiy, Nikolay and Churilov, Ilya},
  title  = {Autocorrelations Decay in Texts and Applicability Limits of Language Models},
  year   = {2023},
  note   = {{arXiv}:2305.06615v1},
  doi = "10.48550/arXiv.2305.06615"
}

@article{miller1955note,
  title={Note on the bias of information estimates},
  author={Miller, George},
  journal={Information theory in psychology: Problems and methods},
  year={1955},
  publisher={Free Press},
}

@article{nakaishi2024statistical,
  title={Statistical properties of probabilistic context-sensitive grammars},
  author={Nakaishi, Kai and Hukushima, Koji},
  journal={Physical Review Research},
  volume={6},
  number={3},
  pages={033216},
  year={2024},
  doi = "10.1103/PhysRevResearch.6.033216",
issn = "2643-1564"
}

@misc{nakaishi2024critical,
  author = {Nakaishi, Kai and Nishikawa, Yoshihiko and Hukushima, Koji},
  title  = {Critical Phase Transition in Large Language Models},
  year   = {2024},
  note   = {{arXiv}:2406.05335v2},
  doi = {10.48550/arXiv.2406.05335},
}

@misc{newville2014lmfit,
  author       = {Newville, Matthew and
                  Stensitzki, Till and
                  Allen, Daniel B. and
                  Ingargiola,  Antonino},
  title        = {LMFIT: Non-Linear Least-Square Minimization and Curve-Fitting for Python
                  },
  month        = sep,
  year         = 2014,
  publisher    = {Zenodo},
  version      = {0.8.0},
  doi          = {10.5281/zenodo.11813},
}

@article{nemenman2001entropy,
  title={Entropy and inference, revisited},
  author={Nemenman, Ilya and Shafee, Fariel and Bialek, William},
  journal={Advances in Neural Information Processing Systems 14},
  year={2001},
}

@techreport{odonnell2009fragment,
    author = "O'Donnell, Timothy J. and Tenenbaum, Joshua B. and Goodman, Noah D.",
    title = "Fragment Grammars: Exploring Computation and Reuse in Language",
    number = "MIT-CSAIL-TR-2009-013",
    institution = "Massachusetts Institute of Technology",
    year = "2009",
url = "https://dspace.mit.edu/handle/1721.1/44963",
note        = {Accessed 2026-03-02},
}

@ARTICLE{sainburg2019parallels,
  title    = "Parallels in the sequential organization of birdsong and human speech",
  author   = "Sainburg, Tim and Theilman, Brad and Thielk, Marvin and Gentner, Timothy Q.",
  journal  = "Nature Communications",
  volume   =  10,
  number   =  1,
  pages    = "3636",
  year     =  2019,
doi = "10.1038/s41467-019-11605-y",
issn = "2041-1723"
}

@article{sainburg2021toward,
  title={Toward a computational neuroethology of vocal communication: from bioacoustics to neurophysiology, emerging tools and future directions},
  author={Sainburg, Tim and Gentner, Timothy Q.},
  journal={Frontiers in Behavioral Neuroscience},
  volume={15},
  pages={811737},
  year={2021},
  doi = "10.3389/fnbeh.2021.811737",
issn = "1662-5153"
}

@article{sainburg2022long,
  title={Long-range sequential dependencies precede complex syntactic production in language acquisition},
  author={Sainburg, Tim and Mai, Anna and Gentner, Timothy Q.},
  journal={Proceedings of the Royal Society B},
  volume={289},
  number={1970},
  pages={20212657},
  year={2022},
doi = "10.1098/rspb.2021.2657",
issn = "1471-2954"
}

@article{schwarz1978estimating,
  title={Estimating the dimension of a model},
  author={Schwarz, Gideon},
  journal={The annals of statistics},
  pages={461--464},
  year={1978},
  publisher={JSTOR},
  doi={10.1214/aos/1176344136},
}

@misc{shen2019mutual,
  title={Mutual information scaling and expressive power of sequence models},
  author={Shen, Huitao},
  note={{arXiv}:1905.04271v1},
  year={2019},
doi = "10.48550/arXiv.1905.04271",
}

@book{stanley1987introduction,
  title={Introduction to Phase Transitions and Critical Phenomena},
  author={Stanley, Harry E.},
publisher = "Oxford University Press",
  year={1971},
edition="",
}

@article{takahashi2017neural,
  title={Do neural nets learn statistical laws behind natural language?},
  author={Takahashi, Shuntaro and Tanaka-Ishii, Kumiko},
  journal={PloS one},
  volume={12},
  number={12},
  pages={e0189326},
  year={2017},
issn = "1932-6203",
doi = "10.1371/journal.pone.0189326"
}

@article{takahashi2019evaluating,
  title={Evaluating computational language models with scaling properties of natural language},
  author={Takahashi, Shuntaro and Tanaka-Ishii, Kumiko},
  journal={Computational Linguistics},
  volume={45},
  number={3},
  pages={481-513},
  year={2019},
  doi = "10.1162/coli_a_00355",
issn = "0891-2017"
}

@ARTICLE{tanaka2016long-range,
  title     = "Long-range memory in literary texts: On the universal clustering of the rare words",
  author    = "Tanaka-Ishii, Kumiko and Bunde, Armin",
  journal   = "PLoS One",
  volume    =  11,
  number    =  11,
  pages     = "e0164658",
  year      =  2016,
issn = "1932-6203",
doi = "10.1371/journal.pone.0164658"
}

@article{tano2020towards,
  title={Towards a more flexible language of thought: Bayesian grammar updates after each concept exposure},
  author={Tano, Pablo and Romano, Sergio and Sigman, Mariano and Salles, Alejo and Figueira, Santiago},
  journal={Physical Review E},
  volume={101},
  number={4},
  pages={042128},
  year={2020},
  doi = "10.1103/PhysRevE.101.042128",
issn = ""
}

@article{wolpert1995estimating,
  title={Estimating functions of probability distributions from a finite set of samples},
  author={Wolpert, David H. and Wolf, David R.},
  journal={Physical Review E},
  volume={52},
  number={6},
  pages={6841},
  year={1995},
  publisher={APS},
  doi = {10.1103/PhysRevE.52.6841},
}

@InProceedings{worth2005growing,
author="Worth, Peter and Stepney, Susan",
title="Growing Music: Musical Interpretations of L-Systems",
booktitle="Applications of Evolutionary Computing",
year="2005",
pages="545-550",
doi       = {10.1007/978-3-540-32003-6_56},
}

@article{youngblood2024language,
  title={Language-like efficiency and structure in house finch song},
  author={Youngblood, Mason},
  journal={Proceedings of the Royal Society B},
  volume={291},
  number={2020},
  pages={20240250},
  year={2024},
doi = "10.1098/rspb.2024.0250",
issn = "1471-2954"
}

@article{zahl1977jackknifing,
  title={Jackknifing an index of diversity},
  author={Zahl, Samuel},
  journal={Ecology},
  volume={58},
  number={4},
  pages={907--913},
  year={1977},
  publisher={Wiley Online Library},
  doi = {https://doi.org/10.2307/1936227},
}

@article{zeldes-etal-2025-erst,
    title = "e{RST}: A Signaled Graph Theory of Discourse Relations and Organization",
    author = "Zeldes, Amir  and
      Aoyama, Tatsuya  and
      Liu, Yang Janet  and
      Peng, Siyao  and
      Das, Debopam  and
      Gessler, Luke",
    journal = "Computational Linguistics",
    volume = "51",
    number = "1",
    month = mar,
    year = "2025",
    address = "Cambridge, MA",
    publisher = "MIT Press",
    doi = "10.1162/coli_a_00538",
    pages = "23--72",
}

@book{zipf2016human,
  title={Human Behavior and the Principle of Least Effort: An Introduction to Human Ecology},
  author={Zipf, George Kingsley},
  year={1949},
  publisher={Addison-Wesley Press}
}

@article{charniak-etal-2000,
  author  = {Charniak, Eugene and Blaheta, Don and Ge, Niyu and Hall, Keith and Hale, John and Johnson, Mark},
  journal = {Linguistic Data Consortium},
  title   = {{BLLIP 1987-89 WSJCorpus Release 1 LDC2000T43}},
  year    = {2000},
doi = {10.35111/fwew-da58}
}

@article{marcus-etal-1993-building,
    title = "Building a Large Annotated Corpus of {E}nglish: The {P}enn {T}reebank",
    author = "Marcus, Mitchell P.  and
      Santorini, Beatrice  and
      Marcinkiewicz, Mary Ann",
    editor = "Hirschberg, Julia",
    journal = "Computational Linguistics",
    volume = "19",
    number = "2",
    year = "1993",
    address = "Cambridge, MA",
    publisher = "MIT Press",
    pages = "313--330"
}

@inproceedings{kitaev-etal-2019,
    title = "Multilingual Constituency Parsing with Self-Attention and Pre-Training",
    author = "Kitaev, Nikita  and
      Cao, Steven  and
      Klein, Dan",
    booktitle = "Proceedings of the 57th Annual Meeting of the Association for Computational Linguistics",
    month = jul,
    year = "2019",
    address = "Florence, Italy",
    publisher = "Association for Computational Linguistics",
    doi = "10.18653/v1/P19-1340",
    pages = "3499--3505",
}

@inproceedings{kitaev-klein-2018,
    title = "Constituency Parsing with a Self-Attentive Encoder",
    author = "Kitaev, Nikita  and
      Klein, Dan",
    booktitle = "Proceedings of the 56th Annual Meeting of the Association for Computational Linguistics (Volume 1: Long Papers)",
    month = jul,
    year = "2018",
    address = "Melbourne, Australia",
    publisher = "Association for Computational Linguistics",
    doi = "10.18653/v1/P18-1249",
    pages = "2676--2686",
}

@inproceedings{merity-etal-2017,
  title={Pointer Sentinel Mixture Models},
  author={Merity, Stephen and Xiong, Caiming and Bradbury, James and Socher, Richard},
  booktitle = "International Conference on Learning Representations",
  year={2017},
}

@inproceedings{bird-loper-2004-nltk,
    title = "{NLTK}: The Natural Language Toolkit",
    author = "Bird, Steven  and
      Loper, Edward",
    booktitle = "Proceedings of the {ACL} Interactive Poster and Demonstration Sessions",
    month = jul,
    year = "2004",
    address = "Barcelona, Spain",
    publisher = "Association for Computational Linguistics",
    pages = "214--217"
}

@article{de-marneffe-etal-2021-universal,
    title = "{U}niversal {D}ependencies",
    author = "de Marneffe, Marie-Catherine  and
      Manning, Christopher D.  and
      Nivre, Joakim  and
      Zeman, Daniel",
    journal = "Computational Linguistics",
    volume = "47",
    number = "2",
    month = jun,
    year = "2021",
    address = "Cambridge, MA",
    publisher = "MIT Press",
    doi = "10.1162/coli_a_00402",
    pages = "255--308",
}

@misc{NPCMJ2016,
  author       = {{National Institute for Japanese Language and Linguistics}},
  title        = {NINJAL Parsed Corpus of Modern Japanese},
  year         = {2016},
  note         = {Version 1.0. Available at \url{https://npcmj.ninjal.ac.jp/interfaces/} (accessed 2025-10-29). Licensed under a Creative Commons Attribution 4.0 International License (\url{https://creativecommons.org/licenses/by/4.0/}).}
}

@article{mann-thompson-1988,
    title = {{Rhetorical Structure Theory}: Toward a functional theory of text organization},
    author = {Mann, William C. and Thompson, Sandra A.},
    pages = {243--281},
    volume = {8},
    number = {3},
    journal = {Text - Interdisciplinary Journal for the Study of Discourse},
    doi = {doi:10.1515/text.1.1988.8.3.243},
    year = {1988}
}

@article{prasad-etal-2014-reflections,
    title = "Reflections on the {P}enn {D}iscourse {T}ree{B}ank, Comparable Corpora, and Complementary Annotation",
    author = "Prasad, Rashmi  and
      Webber, Bonnie  and
      Joshi, Aravind",
    journal = "Computational Linguistics",
    volume = "40",
    number = "4",
    month = dec,
    year = "2014",
    address = "Cambridge, MA",
    publisher = "MIT Press",
    doi = "10.1162/COLI_a_00204",
    pages = "921--950"
}
\bibliographystyle{acl_natbib}

\appendix

\section{Tree Statistics}
\label{app:treestats}

To provide intuition about both the effect of preprocessing on tree shapes and the differences between English and Japanese, we compute basic tree-shape statistics on the original parse trees in BLLIP and NPCMJ. Specifically, we consider the following statistics:
\begin{description}
    \item[Depth.]
    The maximum path length from the root to any node in a tree, averaged over all parse trees.

    \item[Branching Factor.]
    The number of children of an internal node, averaged over all internal nodes.
    This value ranges from $1$ upward and equals $2$ for strictly binary-branching trees without unary nodes.

    \item[Proportion of Unary Nodes.]
    An internal node is \emph{unary} if it has exactly one child.
    We report the proportion of unary nodes, defined as the number of unary internal nodes divided by the total number of internal nodes, excluding POS-to-terminal productions.
\end{description}

As shown in Table~\ref{tab:treestats}, the branching factor and the proportion of unary nodes are comparable between BLLIP and NPCMJ. In contrast, the depth for NPCMJ is smaller than that for BLLIP, suggesting that parse trees in NPCMJ tend to be flatter.

\begin{table*}[t]
    \centering
    \begin{tabular}{lccc}
    \toprule
    Corpus & Depth & Branching & Prop. Unary \\
    \midrule
    BLLIP  & $11.598 (\pm 4.354)$ & $1.527 (\pm 0.904)$ & $0.222$ \\
    NPCMJ  & $6.642 (\pm 3.336)$ & $1.547 (\pm 1.282)$ & $0.271$ \\
    \bottomrule
    \end{tabular}
    \caption{Tree-shape statistics computed on the original parse trees in BLLIP and NPCMJ.
    The depth and the branching factor are reported as mean $\pm$ standard deviation across sentences.}
    \label{tab:treestats}
\end{table*}

\section{Tag Reduction}
\label{app:tag}

To reduce the number of categories, we grouped tags with similar functions into the same category. The eleven categories used for BLLIP and WikiText and the seven categories for NPCMJ are shown in Tables~\ref{tab:natural_compress} and \ref{tab:natural_compress_npcmj}, respectively. 

\begin{table*}[t!]
  \centering \footnotesize
  \begin{tabular}{cl}
      Tags & Part-of-Speech Tags and Phrase Labels \\
      \toprule
      NP & NP NN NNP NNS CD PRP TMP POS QP NNPS EX NX FW \\ \midrule
      VP & VP VBD VB VBN VBG VBZ VBP TO \\ \midrule
      S & S S1 SBAR SINV FRAG SQ SBARQ RRC \\ \midrule
      PP & IN PP RP PRT \\ \midrule
      DT & DT PRP\$ PDT \\ \midrule
      JJ & JJ ADJP JJR JJS NAC ADJ \\ \midrule
      Others & , . '' `` \$ : PRN -RRB- -LRB- \# X INTJ UH SYM LS LST HYPH NFP \\ \midrule
      AUX & AUX MD AUXG \\ \midrule
      RB & RB ADVP RBR RBS \\ \midrule
      CC & CC UCP CONJP \\ \midrule
      WH & WHNP WDT WP WHADVP WRB WHPP WP\$ WHADJP NML \\
      \bottomrule
  \end{tabular}
  \caption{Eleven tags used for BLLIP and WikiText and the corresponding part-of-speech tags and phrase labels.}
  \label{tab:natural_compress}
\end{table*}

\begin{table*}[t!]
  \centering \footnotesize
    \begin{tabular}{cl}
      Tags & Part-of-Speech Tags and Phrase Labels \\ \toprule
      NP & \parbox[t]{0.70\textwidth}{%
      NP NUMCLP PRN NML PNLP NUMCLPSYM N NPR NUM CL ADJN PRO ADJI D Q FN WPRO PNL PRN WNUM WD FW
    } \\ \midrule
      IP & IP VB AX AXD VB0 VB2 PASS PASS2 MD \\ \midrule
      PP & PP P \\ \midrule
      ADVP & ADVP ADV NEG WADV \\ \midrule
      CONJP & CONJP CONJ \\ \midrule
      PP & PP P \\ \midrule
      CP & CP FRAG INTJP FS INTJ \\ \midrule
      Others & LST META LS PU PUL PUR SYM PUQ COMMENT \\
      \bottomrule
  \end{tabular}
  \caption{Seven tags used for NPCMJ and the corresponding part-of-speech tags and phrase labels.}
  \label{tab:natural_compress_npcmj}
\end{table*}

\section{Other Schemes}
\label{app:scheme}

In the main text, we analyzed parse trees after applying two preprocessing steps. First, we binarized the parse trees. Second, we reduced the number of tag types by grouping similar tags, where we use the same reduced categories for both POS and phrasal tags.

Although the statistical quantities depend on the specific preprocessing scheme, we expect that these details do not affect our conclusions, such as whether the MI decays according to a power law or an exponential function, and whether the CFIB is significantly positive.

To examine this expectation, we perform additional analyses on BLLIP under two alternative preprocessing schemes: the unbinarized setting and the phrasal-tag-only setting, in which POS and phrasal tags are distinguished and only the latter are analyzed.

In addition, we analyze unbinarized trees in NPCMJ. Because the original trees in NPCMJ are flatter than those in BLLIP, as shown in Appendix~\ref{app:treestats}, the effect of binarization could be larger.

\begin{figure*}[t!]
    \centering
    \includegraphics[width=\linewidth]{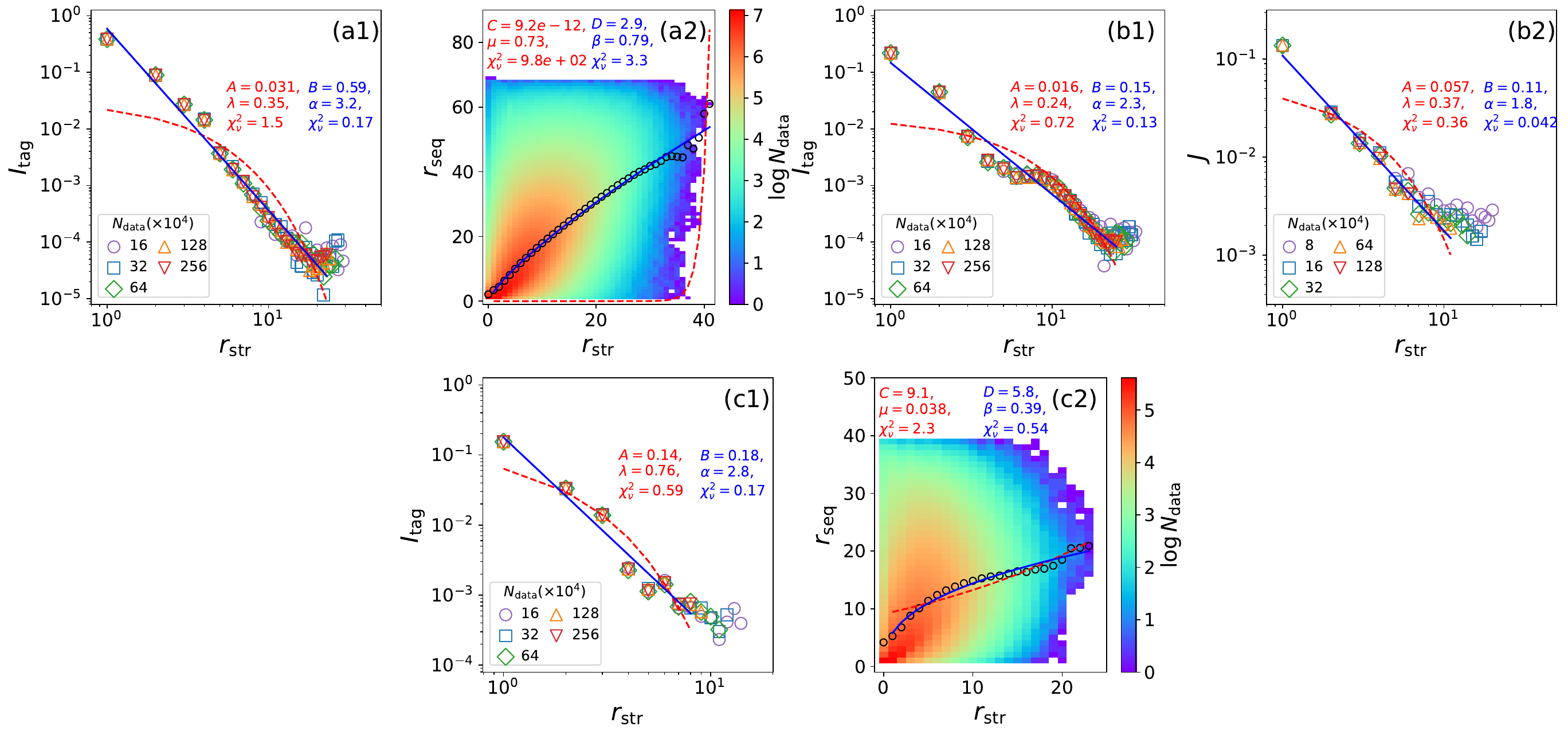}
    \caption{(a1,~2) Results for BLLIP under the unbinarized setting: (a1) MI $I_{\text{tag}}$ between POS or phrasal tags, where fitting was performed for $N_{\text{data}} = 2.56 \times 10^6$; (a2) distribution and average of sequential distances. (b1,~2) Results for BLLIP under the phrasal-tag-only setting: (b1) MI $I_{\text{tag}}$ between phrasal tags, where fitting was performed for $N_{\text{data}} = 2.56 \times 10^6$; (a2) CFIB $J$, where fitting was pwerformed for $N_{\text{data}} = 6.4 \times 10^5$. (c1,~2) Results for NPCMJ under the unbinarized setting: (a1) MI $I_{\text{tag}}$ between POS or phrasal tags, where fitting was performed for $N_{\text{data}} = 2.56 \times 10^6$; (a2) distribution and average of sequential distances.}
    \label{fig:result_other_schemes}
\end{figure*}

Figures~\ref{fig:result_other_schemes}~(a1) and (a2) show the results for BLLIP under the unbinarized setting. Because binarization does not affect the MI in sequences and the dataset is too limited to reliably estimate the CFIB for unbinarized trees, we present only the MI in structures and the sequential distance as a function of the structural distance. For both quantities, the results are qualitatively similar to those for the binarized setting in the main text: the decay of $I_{\text{tag}}$ follows a power-law function rather than an exponential function, and the growth of $r_{\text{seq}}$ is slower than exponential.

In the phrasal-tag-only setting, the MI between POS tags and the sequential distance are not defined. We therefore estimate the MI between phrasal tags and the CFIB, presented in Figures~\ref{fig:result_other_schemes}~(b1) and (b2), respectively. Both quantities exhibit power-law decay, consistent with the results in the main text.

The MI in structures and the sequential distance for unbinarized trees in NPCMJ are presented in Figures~\ref{fig:result_other_schemes}~(c1) and (c2). Again, the results are similar to those in other cases. Both quantities display power-law dependence on the structural distance rather than exponential behavior.

Across all alternative preprocessing schemes, our main conclusion holds: the assumptions in the argument by \citet{lin2017critical} are not satisfied. These results indicate that our findings are robust to the details of preprocessing.

\section{Alternative Estimators}
\label{app:estimators}

In the main text, we computed the MI using the estimator of \citet{grassberger2003entropy} (GR), whose bias decreases rapidly. The estimated values saturate as $N_{\text{data}}$ increases, except for the CFIB in the PCFG. The estimates of the CFIB in the PCFG continue to decrease in the log-scale plot as $N_{\text{data}}$ increases, which is reasonable because the true value is zero. These results show that we have properly estimated the statistical quantities.

To further support the reliability of our estimates, we compute $I_{\text{POS}}$, $I_{\text{tag}}$, and $J$ using six alternative estimators proposed by \citet{chao2003nonparametric} (CS), \citet{horvitz1952generalization} (HT), \cite{miller1955note} and \citet{madow1948limiting} (MM), \citet{nemenman2001entropy} (NSB), \citet{wolpert1995estimating} (WW) and \citet{zahl1977jackknifing} (ZA). For these computations, we use the Python code \texttt{entropy-estimation} implemented by  \citet{arora2022estimating}.

\begin{figure*}[t]
    \centering
    \includegraphics[width=\linewidth]{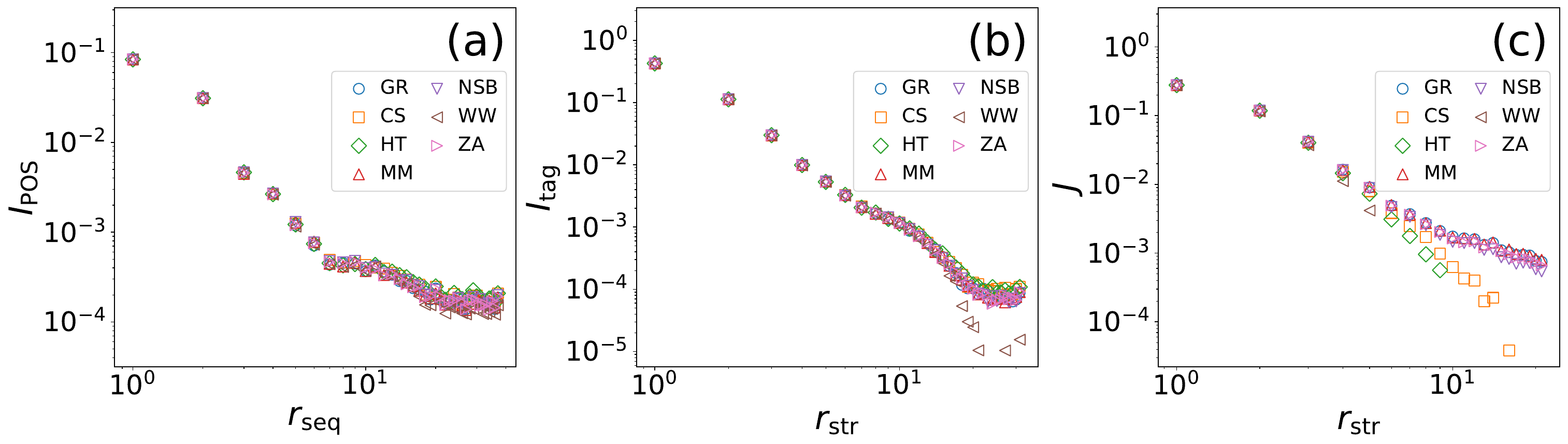}
    \caption{Results for BLLIP obtained with different estimators. (a) MI $I_{\text{POS}}$ between POS tags. (b) MI $I_{\text{tag}}$ between POS or phrasal tags. (c) CFIB $J$ for $(x_0, x_1) = (\text{NP}, \text{NP})$. In all cases, $N_{\text{data}} = 2.56 \times 10^6$.}
    \label{fig:result_estimators_BLLIP}
\end{figure*}

Figure~\ref{fig:result_estimators_BLLIP} presents the results for BLLIP computed with the six alternative estimators, together with those obtained using the GR estimator, which are the same as in the main text. The estimates agree with ours in most cases. Figures~\ref{fig:result_estimators_BLLIP}~(b) and (c) show that the WW estimates for $I_{\text{tag}}$ and $J$ at large distances exhibit clear bias, producing some negative estimates even though the true MI and CFIB are always non-negative. The CS and HT estimates for $J$ also exhibit similar behaviors. These results are consistent with \citet{arora2022estimating}, which reports that these three estimators yield larger errors than the others when the sample size is large. From these results, we can confirm that our estimates in the main text are reliable.

\begin{figure*}[t]
    \centering
    \includegraphics[width=\linewidth]{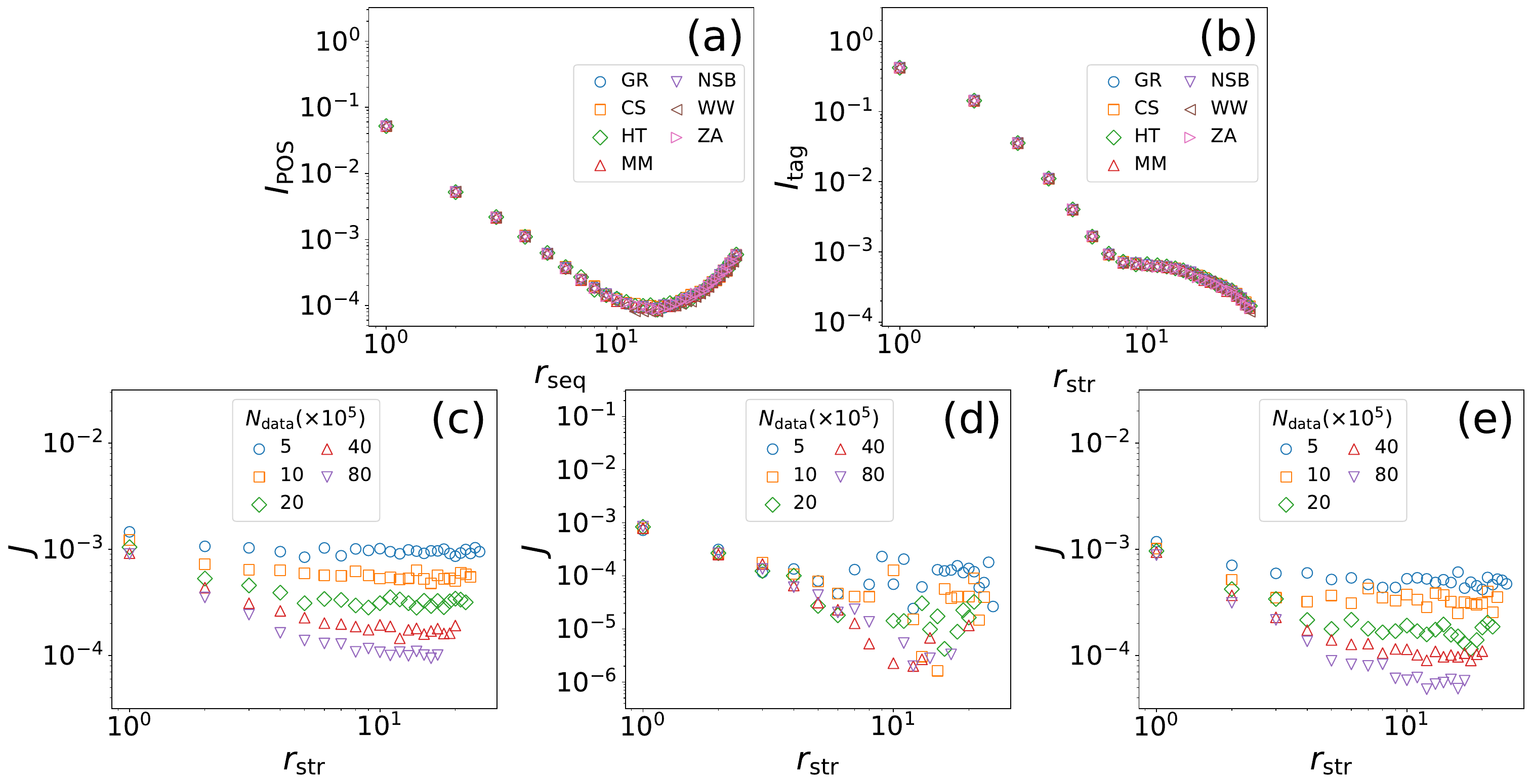}
    \caption{Results for the PCFG obtained with different estimators. (a) MI $I_{\text{POS}}$ between POS tags. (b) MI $I_{\text{tag}}$ between POS or phrasal tags. In both (a) and (b), $N_{\text{data}} = 8 \times 10^6$. (c-e) CFIB $J$ for $(x_0, x_1) = (\text{NP}, \text{NP})$ obtained with (c) MM, and (d) NSB, and (e) ZA.}
    \label{fig:result_estimators_PCFG}
\end{figure*}

Figure~\ref{fig:result_estimators_PCFG} shows the results for the same quantities for the PCFG obtained from the maximum likelihood estimation of BLLIP. The estimates of $I_{\text{POS}}$ and $I_{\text{tag}}$, shown in Figures~\ref{fig:result_estimators_PCFG}~(a) and (b), are consistent across all estimators. This again supports the reliability of the results in the main text.

We also show the estimates of the CFIB $J$ for the PCFG obtained with the MM, NSB, and ZA estimators in Figures~\ref{fig:result_estimators_PCFG}~(c), (d), and (e), respectively. The MM and ZA estimates continue to decrease in the log-scale plot as $N_{\text{data}}$ increases, similar to the GR estimates in the main text. This means that these two estimates properly converge to zero, the true value of the CFIB in PCFGs. The convergence of the NSB estimates is rather unclear. We do not present results with CS, HT, and WW, because most of their estimates are negative and cannot be plotted on a log scale.

In summary, MM, ZA, and GR, which is used in the main text, perform well for all statistical quantities. NSB generally performs well, yet its convergence to the true value is unclear for the CFIB in the PCFG. The other estimators, CS, HT, and WW, sometimes exhibit clear bias, implying that they should be avoided when the sample size is large.

\section{Context-free Independence Breaking}
\label{app:CFIB}

In Section~\ref{subsec:result_CFIB}, we presented the results of CFIB for BLLIP when the two nodes are fixed as $(x_0, x_1) = (\text{NP}, \text{NP})$. In this section, we present the results with $(x_0, x_1) = (\text{NP}, \text{VP})$ and $(\text{VP}, \text{VP})$ to examine how the metric depends on the fixed nodes.

\begin{figure*}[t]
    \centering
    \includegraphics[width=.7\linewidth]{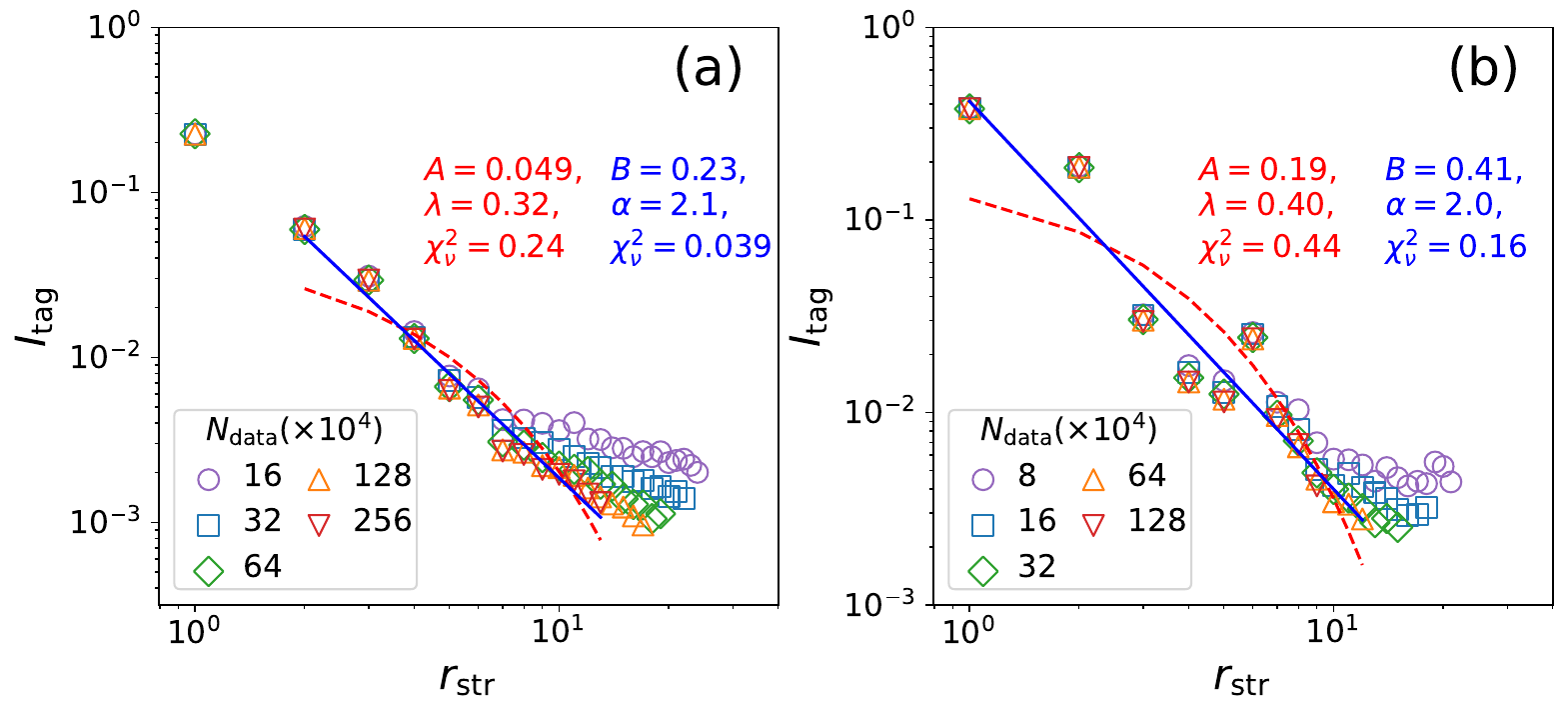}
    \caption{CFIB $J$ for BLLIP with (a) $(x_0, x_1) = (\text{NP}, \text{VP})$ and (b) $(x_0, x_1) = (\text{VP}, \text{VP})$. Fitting was performed for $N_{\text{data}} = 2.56 \times 10^6$ in (a) and $N_{\text{data}} = 6.4 \times 10^5$ in (b).}
    \label{fig:result_CFIB_other_cond}
\end{figure*}

From the results shown in Figure~\ref{fig:result_CFIB_other_cond}, we can observe a power-law decay similar to that with $(x_0, x_1) = (\text{NP}, \text{NP})$, although the power law exponents may be different. This implies that the CFIB in syntactic structures follows a power-law decay, irrespective of the fixed nodes.

\end{document}